\documentclass[10pt,twocolumn,letterpaper]{article}
\usepackage[pagenumbers]{cvpr}
\usepackage[T1]{fontenc}
\usepackage{amsthm}
\usepackage{colortbl}
\usepackage{tikz}
\definecolor{IconGood}{HTML}{2E9E5B}\definecolor{IconBad}{HTML}{E0433B}\definecolor{IconPart}{HTML}{E8A317}\definecolor{IconNone}{HTML}{9AA3AF}
\DeclareRobustCommand{\iconbase}[2]{\tikz[baseline=-.62ex,scale=1.22,line width=.55pt]{\fill[#1] (0,0) circle (.72ex);#2}}
\DeclareRobustCommand{\icok}{\iconbase{IconGood}{\draw[white,line cap=round,line join=round] (-.34ex,-.02ex) -- (-.08ex,-.3ex) -- (.38ex,.28ex);}}
\DeclareRobustCommand{\icno}{\iconbase{IconBad}{\draw[white,line cap=round] (-.28ex,-.28ex) -- (.28ex,.28ex) (-.28ex,.28ex) -- (.28ex,-.28ex);}}
\DeclareRobustCommand{\icpart}{\iconbase{IconPart}{\draw[white,line cap=round] (-.34ex,0) -- (.34ex,0);}}
\DeclareRobustCommand{\icnone}{\iconbase{IconNone}{\draw[white,line cap=round] (-.28ex,-.28ex) -- (.28ex,.28ex) (-.28ex,.28ex) -- (.28ex,-.28ex);}}
\usepackage{algorithm}
\usepackage[noend]{algpseudocode}
\definecolor{Indigo}{HTML}{4F46E5}
\usepackage[pagebackref,breaklinks,colorlinks,allcolors=Indigo]{hyperref}
\usepackage{preprint}
\def\confName{CVPR}
\def\confYear{2027}

\newtheorem{proposition}{Proposition}
\newcommand{\supp}[1]{App.~#1}
\title{Was the Model Given the Right Action?\\Auditing and Repairing Action Interfaces in Navigation World Models}
\ppauthors{Ajmain Istiak Apon\textsuperscript{1},\quad Md Mehedi Hasan\textsuperscript{1},\quad Sejuti Rahman\textsuperscript{2}}
\ppaffils{\textsuperscript{1}University of Dhaka\qquad \textsuperscript{2}New Uzbekistan University}
\ppcontact{}
\pplinks{}
\newcommand{\topic}[1]{\par\smallskip\noindent{\sffamily\bfseries\color{PPInk}#1}\enspace\ignorespaces}
\ppabstract{World models are evaluated by predicting future observations from actions. These evaluations assume that actions are prepared the same way as during training, but an unchecked mismatch can make a model appear less accurate than it is. We audit this assumption in Navigation World Models (NWM) and find three interacting differences between its training and rollout code. Correcting the inputs, without retraining, reduces mean visual prediction error by 42\%, measured with DreamSim on the RECON navigation dataset. The correction also changes the ranking of NWM checkpoints in an exploratory analysis and removes or reduces two alternative models' measured advantages. Correcting only the normalization offset instead increases error, showing why interacting mismatches must be checked together. To test whether sensitivity to input preparation extends beyond NWM, we deliberately introduce encoding errors into two alternative models; both become less accurate. We introduce ActRosetta, a tool that checks whether action encoding matches training, identifies supported causes of disagreement, and repairs inputs when sufficient information is available. Repair helps little where predictions barely respond to actions, the tool's diagnosis has limited coverage, and improved control remains unproven. For any action-conditioned world model, the lesson is the same: before reporting rollout error, verify that evaluation encodes actions exactly as training did.}
\begin{document}
\raggedbottom
\hfuzz=0.1pt
\twocolumn[{%
\renewcommand\twocolumn[1][]{#1}%
\maketitle
\begin{center}
\includegraphics[width=\textwidth]{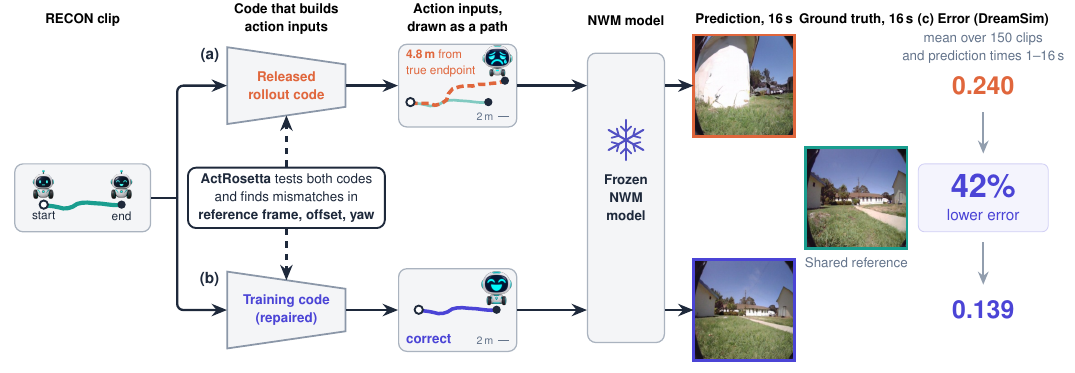}
\captionof{figure}{\textbf{Does NWM receive the action inputs it learned?} NWM~\cite{bar2025navigation} uses separate training and rollout code that encode the same robot motion differently. On one RECON clip, we run the frozen checkpoint with (a) \emph{released actions}, built by its released rollout code, and (b) \emph{repaired actions}, built from the same poses by its training code. By 16\,s, (a) has drifted from ground truth while (b) stays close. Only the action inputs change. (c) Over 150 clips and prediction times 1--16\,s, repaired actions reduce mean DreamSim error by 42\%. ActRosetta (dashed arrows) queries both code paths on test paths to identify encoding mismatches.}
\label{fig:method}
\end{center}
\vspace{.6em}
}]

\section{Introduction}
World models are evaluated through \emph{rollouts}: starting from real camera frames of a recorded robot run, or \emph{trajectory}, the model receives the run's actions and predicts the frames that follow, each prediction feeding the next. Evaluations use segments of trajectories, called \emph{clips}. We study Navigation World Models (NWM)~\cite{bar2025navigation}, a model that predicts future camera frames from past frames and navigation actions; later work reuses its rollout code, so one error can affect several evaluations. Preprocessing maps poses to action inputs, choosing a reference frame, a normalization offset, an angle convention and a rule for aggregating raw-step displacements. This \emph{action encoder} has no learned weights: it is neither NWM's image encoder nor its learned action embedding. A difference between the encodings used in training and in evaluation is a \emph{mismatch}; the error it causes is charged to the model.

Existing benchmarks ask whether a model follows the action it is given~\cite{yang2026mirabench,chen2026robotic,co2026worldsimprobe}; we ask whether the model received the action it learned to interpret. Our main result is an audit of NWM's action interface: its released rollout code prepares actions differently from its training code, through three interacting mismatches that change one frozen checkpoint's predictions (Figure~\ref{fig:method}). We call the action inputs built by the released rollout code \emph{released actions}, and those built from the same poses by the training code \emph{repaired actions}; replacing one with the other, with the checkpoint unchanged, is \emph{repair}. A \emph{chunk} spans one prediction call, which predicts one future state a fixed time shift ahead. The two encodings agree only on the first chunk, so a one-chunk check passes.

Detecting that two action encoders differ is differential testing; knowing what to fix is harder, because mismatches interact: fixing only NWM's offset increases error. The encoding also spans the rollout loop, dataset code and normalization, so reading one function can miss what the model receives. We therefore query action encoders, with a tool we call ActRosetta. Our NWM repair itself used NWM's own training code; ActRosetta lets others run the same check on their own encoders. Our contributions are:
\begin{enumerate}
\item \textbf{An action-interface mismatch that changes model evaluation.} NWM's rollout code differs from training in reference frame, normalization offset and yaw wrapping. Repair, without retraining, reduces mean DreamSim error on RECON from 0.240 to 0.139 (42\%) and removes or shrinks two alternative models' leads (Sec.~\ref{sec:models}).
\item \textbf{A repeatable audit with witnessed attribution.} ActRosetta queries paired training and evaluation encoders on controlled paths, names a differing field only when a test path witnesses it, and repairs only when the whole contract, an encoder's full set of encoding choices, is determined. Unlike ActionABI~\cite{attri2026actionabi}, which shares its principles but recovers contracts from logs, it diagnoses paired code, identifying the interacting choices a patch must address (Secs.~\ref{sec:contract}--\ref{sec:aicr}).
\item \textbf{Evidence separating detection, attribution and repair.} ActionFaultBench, our benchmark of action-encoder pairs whose differences are known in advance, tests whether a checker can find differences, explain them and recover the training encoder's inputs, on four sealed test sets written by separate automated agents after our code was frozen; an audit of 40 public repositories complements it. On the final 240-pair test, ActRosetta detects all 91 differing pairs, recovers their differences exactly in 73 cases, and repairs all 38 differing pairs supported by output conversion (Secs.~\ref{sec:bench}--\ref{sec:results}).
\end{enumerate}

\section{Related Work}
\begin{table}[t]
\centering\footnotesize
\setlength{\tabcolsep}{3pt}
\caption{The closest related work. The first five evaluate what a model or controller does with the actions it receives; ActionABI and ActRosetta examine how actions are encoded before they reach the model.}
\label{tab:related}
\begin{tabular}{>{\raggedright\arraybackslash}p{.29\linewidth}>{\raggedright\arraybackslash}p{.27\linewidth}>{\raggedright\arraybackslash}p{.34\linewidth}}
\toprule
Work & Evidence & Question\\
\midrule
MiraBench~\cite{yang2026mirabench} & Model rollouts & Are actions followed?\\
WorldEcho~\cite{chen2026robotic} & Off-expert actions & Are actions followed?\\
WorldSimProbe~\cite{co2026worldsimprobe} & Controlled action changes & Is the model faithful?\\
WorldOracle~\cite{duan2026worldoracle} & Paired real executions & Is the rollout faithful?\\
Mind the Execution Gap~\cite{wen2026mind} & Execution timing & Are actions applied on time?\\
ActionABI~\cite{attri2026actionabi} & State--action logs & Which encoding was used?\\
ActRosetta (ours) & Training and evaluation encoders & Do they match, and if not, why?\\
\bottomrule
\end{tabular}

\end{table}

\topic{Navigation world models.}
Later work builds on NWM~\cite{bar2025navigation} to improve rollout consistency~\cite{yan2026mwm}, use different visual representations~\cite{zhang2026rae}, generate interactive video~\cite{huang2025vid2world}, plan with compact representations~\cite{kim2026planning} and model long histories efficiently~\cite{guo2026visual}. World-in-World places NWM inside closed-loop tasks, where its predictions guide an agent~\cite{zhang2026world}. Such work can reuse NWM's rollout code: in our audit, MWM's executed rollout shares all three mismatches (Sec.~\ref{sec:results}).

\topic{Do world models follow actions?}
Several recent benchmarks test whether a model's predictions respond correctly to the actions it is given (Table~\ref{tab:related}). MiraBench finds that visual quality is a poor guide to whether actions are followed~\cite{yang2026mirabench}; WorldEcho shows that models can fail on action sequences unlike those an expert would choose~\cite{chen2026robotic}; WorldSimProbe states what a simulator's outputs should show when actions are changed in a controlled way~\cite{co2026worldsimprobe}; WorldOracle compares predictions with paired executions of the same actions in the real world~\cite{duan2026worldoracle}; and ActSWM finds that some models' predictions become nearly identical under different action sequences~\cite{gan2026actswm}. These works judge what the model does with an action, which complements our check: a model can follow actions well and still be judged unfairly if its inputs are encoded differently.

\topic{Mismatched action inputs.}
Mind the Execution Gap~\cite{wen2026mind} addresses mismatches in time, such as delay, packet loss and reordering when a frozen controller is executed asynchronously, and corrects them with an interface designed for each architecture; our mismatches are in the encoded values themselves. ActionABI~\cite{attri2026actionabi} is closest in method. From logged state--action trajectories, it recovers the conventions an action encoder follows, keeps every candidate that fits the logs equally well, and converts actions only when the candidates agree, abstaining otherwise. ActRosetta follows the same principles but differs in evidence and goal: instead of logs from one system, it runs a model's training and evaluation encoders on chosen test paths, explains where they disagree, and repairs the evaluation inputs. We do not claim that it recovers conventions better than ActionABI.

\topic{Evaluation details that change results.}
Experimental choices and run-to-run variability can change model comparisons~\cite{henderson2018deep,bouthillier2021accounting}, ignoring statistical uncertainty can change conclusions~\cite{agarwal2021deep}, and data leakage can inflate performance~\cite{kapoor2022leakage}. Simple baselines have shown that single-view reconstruction~\cite{tatarchenko2019single} and absolute pose regression~\cite{sattler2019understanding} networks do something different from what their scores suggested. Closest to our setting, how images are resized and compressed changes the metrics used to evaluate GANs~\cite{parmar2022aliased}, and rotation representation affects how well networks learn~\cite{zhou2019continuity}; we study the same kind of step for action inputs, including how yaw is represented.

\section{How Action Encoding Changes NWM's Evaluation}
\label{sec:models}
We measure how released and repaired actions change NWM's error, motion and comparisons, why partial corrections fail, and where correction matters.

\topic{The three mismatches.}
NWM's training and rollout encoders differ in three ways.\footnote{Repository \url{https://github.com/facebookresearch/nwm}, commit \texttt{3f6cd8e}, accessed October 2026. Training actions: \texttt{TrainingDataset} (\texttt{datasets.py}); rollout actions: \texttt{EvalDataset} (\texttt{datasets.py}) and \texttt{generate\_rollout} (\texttt{isolated\_nwm\_infer.py}).}
First, training expresses each chunk's motion in the robot's reference frame at that chunk's start; the rollout code keeps the reference frame of the first chunk. Second, training normalizes each displacement $(x,y)$, in units of the dataset's waypoint spacing, to $(0.267x-1/3,\;0.25y)$ (\texttt{normalize\_data} in \texttt{misc.py}, with bounds from \texttt{config/data\_config.yaml}). The constant $b=(-1/3,0,0)$ on $(x,y,\text{yaw})$ is the \emph{normalization offset}; the NWM paper does not mention it. Training adds it to every chunk; the rollout code adds it only to the first. Third, training wraps each chunk's yaw (heading change) to $[-\pi,\pi)$; the rollout code differences yaw angles wrapped relative to the start, so a chunk's yaw jumps by about $2\pi$ when the cumulative turn crosses $\pm\pi$.

\topic{How we measure the effect.}
We compare predicted with recorded images using DreamSim~\cite{fu2023dreamsim}, a learned perceptual distance, and LPIPS~\cite{zhang2018unreasonable}; lower is better. Paired runs share clips and diffusion noise, so only the action inputs differ. RECON~\cite{shah2021rapid} (outdoor, in NWM's training data) is the main dataset; SCAND~\cite{karnan2022socially} (also in training) and Go Stanford~\cite{hirose2018gonet} (indoor, not in training) are additional tests. The \emph{main test} uses NWM's own 150 RECON rollout clips of 16\,s (20 trajectories), its largest public checkpoint (XL), 250 diffusion steps as in NWM's evaluation, seed 0, and 1 predicted frame per second (fps). A \emph{new cohort} of 120 clips from 60 trajectories, recorded on 20 days and not examined before, tests all five public checkpoints~\cite{bar2025navigation}: sizes S, B, L and XL, and XL trained with additional Ego4D video (three seeds). Brackets give 95\% trajectory-bootstrap confidence intervals~\cite{efron1992bootstrap} unless stated otherwise. A \emph{locked} test had its protocol and analysis fixed before its results were seen (recorded locally, not publicly registered); exploratory work preceded the locked tests. Settings and further intervals are in \supp{M}.

\begin{figure*}[t]
\centering\includegraphics[width=\textwidth]{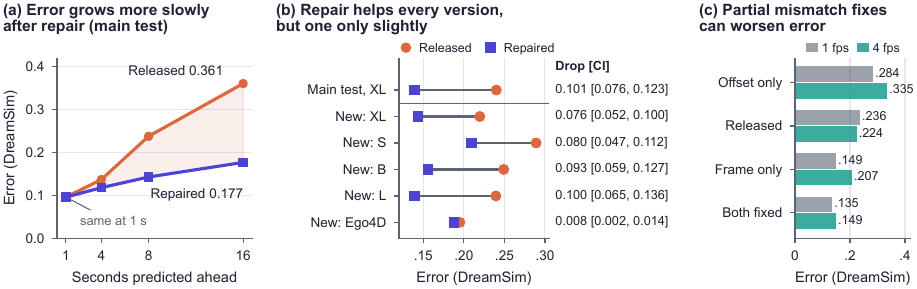}
\caption{\textbf{Repairing action inputs reduces NWM's rollout error.} (a) Main test (150 RECON clips, 250 diffusion steps): mean DreamSim error at each prediction time. (b) Mean error over 1--16\,s and its drop, with 95\% trajectory-level intervals for the main test and 99\% day-level intervals, corrected for five tests, for the five public checkpoints on the new cohort (120 clips, three seeds). (c) Correcting one mismatch at a time (50 main-test clips, 50 diffusion steps): correcting only the offset increases error; full repair (``Both fixed'') is lowest.}
\label{fig:rollout}
\end{figure*}
\topic{Repair reduces rollout error.}
In the main test, repair lowers mean DreamSim error, averaged over clips and prediction times 1--16\,s, from 0.240 to 0.139, a reduction of 0.101 [0.076, 0.123] (42\%), without retraining. At 16\,s the reduction is 0.185 [0.151, 0.212] (51\%; Figure~\ref{fig:rollout}a,b). It holds across noise seeds and at 4\,fps, and enabling NWM's \texttt{torch.compile}, which our generation code omits, leaves the 16\,s reduction almost unchanged. With NWM's own evaluation script, our released-action LPIPS at 8 and 16\,s is within 5\% of NWM's published values, and repaired-action LPIPS is 20\% and 25\% below them; as the released weights differ from the paper's and we reran one sample rather than five, this is consistent with, but does not prove, the published results having used the released encoding (\supp{E}). On the new cohort, all five checkpoints improve, four by more than 0.02, the smallest useful reduction we had set in advance. The Ego4D-trained checkpoint gains only 0.008, for reasons we cannot establish; with no Ego4D dataset class in the public code, its training encoding cannot be verified. These are variants of one model family, so this test alone does not establish generality.

\topic{Repair also improves motion.}
Lower image error need not mean better action following. In a locked test on 50 main-test clips (15 trajectories, 50 diffusion steps), each clip's own actions and four other trajectories' actions generate five 16\,s futures; DreamSim picks the own-action future as closest to the recorded 16\,s image for 15 of 50 clips with released actions and 40 of 50 with repaired actions (trajectory-averaged accuracy 0.50 [0.31, 0.69] and 0.86 [0.71, 0.97]; chance 0.20; LPIPS accuracy gain 0.43 [0.08, 0.73]). An inverse dynamics model, exploratory because it failed its locked forward-motion check (\supp{F}), agrees: paths it estimates from repaired predictions are 1.39\,m [0.89, 1.87] closer to the recorded path than those from released predictions (mean over 1--16\,s and 150 clips), and released predictions are not measurably closer to the path their actions describe under the training encoding.

\begin{figure}[t]
\centering\includegraphics[width=\linewidth]{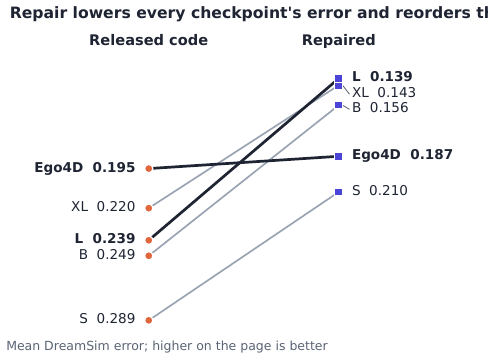}
\caption{\textbf{Repair changes the order of NWM's checkpoints.} Each line connects one checkpoint's mean DreamSim error with released actions (left) and repaired actions (right) on the new cohort of Figure~\ref{fig:rollout}b; higher means lower error. Exploratory.}
\label{fig:ranking}
\end{figure}

\topic{The encoding changes which checkpoint looks best.}
On the new cohort, the Ego4D-trained checkpoint moves from first to fourth after repair, and L becomes best (Figure~\ref{fig:ranking}). Four of the ten pairwise orderings reverse under both DreamSim and LPIPS, with day-level confidence intervals excluding zero and all four significant after Holm correction for ten comparisons; these placements hold when any one of the 20 recording days is left out. The analysis is exploratory because we chose it after seeing the means; the Ego4D checkpoint also differs in training data and length.

\begin{figure}[t]
\centering\includegraphics[width=\linewidth]{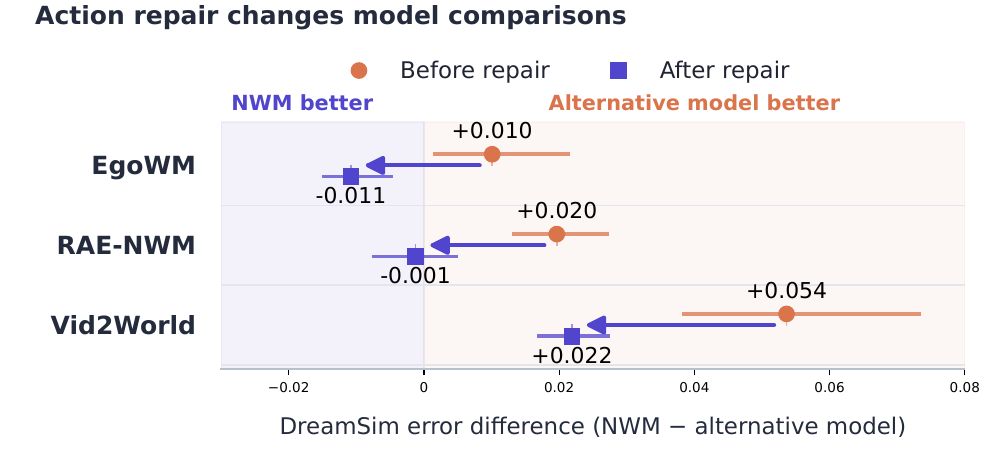}
\caption{\textbf{Repairing NWM's actions removes or shrinks the alternative models' leads.} Arrows go from NWM's released actions (circles, ``before repair'') to its repaired actions (squares, ``after repair''). The axis is NWM's DreamSim error minus the alternative model's; left of zero favors NWM. Bars: 95\% trajectory-level intervals. EgoWM, RAE-NWM: 90 clips of 4\,s, five seeds; Vid2World: 30 clips, one seed.}
\label{fig:cross}
\end{figure}

\topic{The encoding affects comparisons between models.}
We then compare NWM with three alternative models: EgoWM~\cite{bagchi2026walk}, which adds action conditioning to pretrained video diffusion models; RAE-NWM~\cite{zhang2026rae}, which predicts navigation futures in a dense visual representation space; and Vid2World~\cite{huang2025vid2world}, which adapts video diffusion for interactive prediction. Each runs at its own settings on the same 4\,s RECON clips (16 frames at 4\,fps), under a comparison rule locked in advance; only NWM's action inputs change (Figure~\ref{fig:cross}). On 90 clips (19 trajectories, five seeds, one GPU type), EgoWM and RAE-NWM have lower DreamSim error than NWM with released actions, by 0.010 [0.002, 0.021] and 0.020 [0.013, 0.027]. After repair, NWM has lower error than EgoWM by 0.011 [0.005, 0.015], and RAE-NWM's lead becomes $-$0.001 [$-$0.007, +0.005]: no detectable difference, which does not establish equivalence. Under LPIPS, neither repaired difference excludes zero. Thirty of the 90 clips come from an earlier, smaller run; on the other 60, EgoWM has no clear lead even with released actions. Vid2World, run only in that earlier run (30 clips, one seed), keeps a reduced lead: 0.022 [0.017, 0.027], from 0.054 [0.038, 0.073]. EgoWM's training code is not public, so its action interface is unverified, and the results describe our matched comparisons, not published scores.

\topic{Mismatches interact, so partial corrections can hurt.}
Correcting one mismatch at a time instead of all three (50 main-test clips, 15 trajectories, 50 diffusion steps, seed 0; Figure~\ref{fig:rollout}c; exploratory), we find that correcting only the normalization offset increases DreamSim error over released actions, by 0.048 [0.028, 0.062] at 1\,fps and 0.110 [0.054, 0.139] at 4\,fps. Correcting only the reference frame leaves 0.014 [0.005, 0.023] more error than full repair at 1\,fps and is not measurably better than released actions at 4\,fps ($-$0.017 [$-$0.039, +0.005]); correcting only yaw changes little. Conversely, adding only the frame mismatch to repaired actions raises error by 0.150 [0.107, 0.176], more than all three together, while adding only the offset mismatch does not measurably hurt (0.006 [$-$0.002, +0.015]), so correcting only the offset leaves the most damaging mismatch in place. The whole encoding must therefore be checked before a patch is applied.

\topic{Injected faults also hurt the alternative models.}
We also plant encoding errors ourselves, called \emph{injected faults}; \emph{mismatch} stays reserved for differences in a model's own code. In NWM, adding $\pm2\pi$ to the yaw of every chunk after the first keeps the path but puts every yaw input far outside the training range, and raises error over repaired actions by 0.128 [0.106, 0.162] (exploratory), as much as released actions do, so a check must compare numerical inputs, not only the paths they describe. All 14 injected faults increase NWM's error, from 0.006 (one-step lag) to 0.513 (reversed sign), and ActRosetta recovers the repaired actions exactly for all 14. To test whether this sensitivity extends beyond NWM, we inject frame, offset, combined (frame plus offset, as no clip has a natural yaw jump within 4\,s) and $2\pi$ faults into EgoWM and RAE-NWM on the 90 comparison clips, adapting the offset and $2\pi$ faults to RAE-NWM's convention; none is a bug in either model's released code (\supp{T}). Every fault hurts both models. The combined fault raises DreamSim error by 0.029 [0.024, 0.035] in EgoWM and 0.033 [0.026, 0.040] in RAE-NWM (seed 0, RTX 3090 Ti) and by 0.027 and 0.035 (seeds 1--2, A100 GPUs), all eight passing Holm correction within each GPU type; the path-preserving $2\pi$ fault hurts most, by 0.180 and 0.092.

\topic{Where repair matters.}
The effect also depends on how a model represents actions and how strongly its predictions respond to them. Removing the normalization offset raises Vid2World's DreamSim error by 0.059 [0.039, 0.083] (30 clips, one seed), whereas injected faults do not clearly hurt DrivingWorld~\cite{hu2026drivingworld}, a driving world model that takes actions as discrete tokens, and doubling its commanded distance even lowers its error by 0.037 [0.020, 0.059] (12 demonstration clips, one seed); lower prediction error alone therefore cannot establish that an encoding is correct. The \emph{action-responsiveness ratio} divides the image change caused by substituting another clip's actions by the change caused by a new random seed; above one, actions move predictions more than noise does. Encoding changes have clear effects where it is high, in NWM on RECON (2.25) and in Vid2World (2.54), and small ones where it is near one, in NWM on SCAND (1.10) and Go Stanford (0.90), where repair lowers error by only 0.006 and 0.005, and in DrivingWorld (0.995); injected faults hurt EgoWM (2.41) and RAE-NWM (1.84). This pattern is descriptive: the tests differ in clips, frame rates and horizons. Better images need not mean better decisions: in Habitat~\cite{savva2019habitat} with Matterport3D~\cite{chang2017matterport3d} scenes, repair does not measurably improve command following (40 cases) or image-goal decisions (20 cases, two runs; \supp{Q}), so we make no claim of improved control. NWM's own planning code builds actions by a separate route that matches training at default settings, so these mismatches do not reach its planning results.

\section{Action-Interface Contracts}
\label{sec:contract}
To compare encoders, we describe each one's choices by an \emph{action-interface contract}, or \emph{contract}: which time steps are used, which reference frame expresses motion, and how values are scaled, shifted and represented. A contract describes preprocessing, not the world model's learned behaviour. At raw time step $t$, the robot's pose is $g_t=(p_t,\theta_t)$, with position $p_t\in\mathbb R^2$ and heading $\theta_t$. Given a recorded path $g_{0:N}$ and a chunk size $s$ in raw time steps, an action encoder $E$ returns one action row of $d$ numbers per chunk, $K$ rows in all: a tensor in $\mathbb R^{K\times d}$.

\topic{The contract family.}
We write $\phi$ for a contract and $E_\phi$ for the encoder it defines. The \emph{contract family}, or \emph{family}, is the set of contracts our method can represent, not every possible implementation. Chunk $k$ reads from $i_k=kms-\ell$ to $j_k=i_k+ms$, where the lag $\ell$ shifts the read window and the stride multiple $m$ scales its length (default $\ell=0$, $m=1$). The action row has a position part and a yaw part:
\begin{equation}
\begin{aligned}
E_\phi(g)_k=\big[\,&\mathrm{sat}_C\big(\mathcal A\big(R(\alpha_k)^{\top}(p_{j_k}-p_{\kappa_k})\big)M\\
&\quad+\mathbf 1[k\in\Omega]\,b\big),\\
&Q\big(\lambda_\theta\mathcal A_\theta W(\theta_{j'_k},\theta_{\eta_k})+b_\theta\big)\,\big].
\end{aligned}
\label{eq:family}
\end{equation}
In the position part, the anchor index $\kappa_k$ selects the chunk start or sequence start as origin; $R(\alpha_k)^\top$ expresses displacement in the reference frame of heading $\alpha_k$ (current, first, or zero for world coordinates); the rule $\mathcal A$ takes the sum, mean or final raw-step displacement; the $2\times2$ matrix $M$ covers scale, axis swaps and sign changes; the normalization offset $b$ is added on the chunks in $\Omega$ (every chunk or only the first); and $\mathrm{sat}_C$ limits each component to a saturation bound $C$, if one is present. In the yaw (heading-change) part, the reference index $\eta_k$ selects the chunk start, sequence start or zero; the rule $W$ takes a wrapped change, an unwrapped change, or a difference of separately wrapped headings (wrapping restricts an angle to a fixed range, so a whole turn may be discarded); the endpoint $j'_k$ has its own timing; and after aggregation $\mathcal A_\theta$, scale $\lambda_\theta$ and offset $b_\theta$, $Q$ represents the result as a scalar, a sine--cosine pair or a cosine--sine pair. The discrete choices give 720 position and 420 yaw hypotheses; continuous parameters are fitted from encoder outputs. NWM's training and rollout encoders are two contracts in this family that differ in $\alpha_k$, $\Omega$ and $W$: the three mismatches of Sec.~\ref{sec:models}.

\topic{What can be recovered from queries.}
We call an encoder on test paths, called \emph{probes}, and observe its outputs without reading its code. Different contracts may produce exactly the same outputs, so our target is a \emph{behavioural class}: all contracts that agree on every allowed input. The allowed inputs form the \emph{admissible domain} $\mathcal D$: by default, all planar pose sequences with chunk sizes $s\in\{1,2,4\}$, including turns that cross the angle-wrap boundary within a chunk; the implementation compares contracts on $\mathcal D$ using a fixed finite set of long paths. Each class is reported by its \emph{canonical} member, the one with the fewest non-default discrete choices, as a convention, not evidence about the source code. A field is \emph{unexcited} when the probes do not vary the motion needed to reveal it (a path with no rotation may not distinguish heading conventions), and an encoder is \emph{out of family} when no contract in the family reproduces its outputs within tolerance. Standing still reveals an offset, translations reveal axis mappings, and turns reveal reference-frame and yaw conventions. Proposition~\ref{prop:id} formalizes this for a restricted \emph{core family}, without timing shifts or saturation.

\begin{proposition}[Identifiability up to equivalence]
\label{prop:id}
Take deterministic encoders in the core family (no saturation, lag or stride multiple) with chunk size $s\in\{1,2,4\}$, invertible $M$, nonzero yaw scale, frame in \{world, first, current\}, offset on all chunks or the first, and yaw rule in \{wrapped, unwrapped, difference of wrapped\}. A probe set with zero motion, two non-collinear translations, a nonzero start heading, a turn followed by a translation, and a chunk rotating by more than $\pi$ determines the behavioural class on $\mathcal D$.
\end{proposition}
\supp{J} gives its proof sketch and the weaker evidence for the full family, where lag, stride and aggregation interact: an exhaustive finite check at one setting and empirical support.

\begin{algorithm}[t]
\caption{Compare two encoders and repair}
\label{alg:compare}
\footnotesize
\begin{algorithmic}[1]
\Require accepted sets $H_T,H_E$; encoders $E_T,E_E$; outputs $o$; poses if given
\If{$H_T$ or $H_E$ was refused}
  \State no field claim; report only if $E_T,E_E$ differ on fixed paths
  \State \Return
\EndIf
\For{each field $f$}
  \State \textbf{mismatch} if $f$ differs for every pair in $H_T\times H_E$
  \State \textbf{agreement} if $f$ is equal for every pair
  \State \textbf{unresolved} otherwise, or if unexcited on either side
\EndFor
\State saturation: unresolved if one side never passed the bound
\State check any \textbf{agreement} against $E_T,E_E$ directly
\If{outputs differ but no field explains it} \State report \emph{differs, unattributed} \EndIf
\State $\mathcal S\gets$ smallest field sets that turn $\phi_E$ into $\phi_T$
\For{each mismatch $f$ in every set of $\mathcal S$}
  \State search paths for a witness of $f$ (Sec.~\ref{sec:aicr})
  \State name $f$ if a witness exists, else mark it unresolved
\EndFor
\If{poses given and $H_T$ definite} \Return $E_{\phi_T}(\text{poses})$
\ElsIf{both definite, $\phi_E$ invertible} \Return $E_{\phi_T}\!\circ\! E_{\phi_E}^{-1}(o)$
\Else \ abstain, and request the named probe or the poses \EndIf
\end{algorithmic}
\end{algorithm}

\section{ActRosetta}
\label{sec:aicr}
ActRosetta, our tool, calls a training and an evaluation encoder on the same probes and answers at two levels. The first, the \emph{family-free check}, is differential testing, which we do not claim as new: it reports whether their outputs differ on a fixed path set, with a path where they do (Algorithm~\ref{alg:compare}, lines~2 and~9); Sec.~\ref{sec:results} compares it with random fuzzing. The second level, which recovers both contracts, is what ActRosetta adds to detection: it names the fields a patch must change, since mismatches interact, and repairs only when the whole contract is determined.

\topic{Probes.}
Seven kinds of probe (the probe set $B$), each of at most 20 \emph{transitions} (moves between consecutive poses), expose particular choices: standing still, motion along each axis, rotation in place, turning then moving, crossing the wrap, several chunk sizes, and single impulses. Three long, fast \emph{range paths} $L$ test the fit at high speed, and three random hold-out paths $V$ verify it, at tolerance $\epsilon$.

\topic{Recovery, refusal and abstention.}
For each encoder, recovery fits every contract in the family to its probe outputs and \emph{accepts} those that fit within $\epsilon$ and pass verification on $L$ and $V$. For training and evaluation encoders $E_T$ and $E_E$, with contracts $\phi_T$ and $\phi_E$, Algorithm~\ref{alg:compare} takes the accepted sets $H_T$ and $H_E$ and calls a field a mismatch only if it differs for every pair of accepted contracts, including contracts that behave alike. ActRosetta \emph{refuses} an encoder that is nondeterministic, outside the family or failing verification, or whose accepted contracts miss fresh, faster paths; it then makes no field claim and reports only the family-free check (line~2). It \emph{abstains} when it leaves a field unresolved or declines to repair (line~18).

\topic{Witness-backed naming.}
Two contracts can differ in a field that does not explain their difference. ActRosetta therefore changes the evaluation contract toward the training one, field by field, to find the smallest field sets that make their outputs agree, and names a field only if it is in all of those sets and has a \emph{witness}: a path on which both real encoders match their contracts while undoing every other change still leaves an output gap of at least $10^{-3}$ of the output scale. We call this rule \emph{witness-backed naming}; other fields stay unresolved, and some are never named separately. If each encoder's true contract is accepted, every reported mismatch is a true difference, since the true pair is among those examined. The premise fails for an encoder that departs only beyond every tested path, which no finite black-box test can exclude, so we state claims only on the tested range, $\mathcal D_{\text{valid}}=\{g_{0:N}:\|\Delta p_t\|\leq 2.7\text{\,m},\ |\Delta\theta_t|\leq 1.05\text{\,rad},\ s\in\{1,2,4\}\}$, as tested evidence, not proof.

\topic{Repair.}
Given poses, the repaired input is $E_{\phi_T}(g)$, which equals calling the training encoder; this is how we repaired NWM in Sec.~\ref{sec:models}, so that result does not depend on ActRosetta. Given only evaluation outputs, ActRosetta applies the \emph{adapter} $E_{\phi_T}\circ E_{\phi_E}^{-1}$, which maps evaluation outputs to training-consistent inputs. This requires \emph{invertibility with respect to $\phi_T$}: the evaluation outputs must determine the chunk-boundary poses read by $\phi_T$, which fails under lag, stride multiples, last-step aggregation and saturation, and under wrapped evaluation yaw when training requires unwrapped yaw. No public evaluation we found needs the adapter; we tested it on ActionFaultBench and on 20 RECON clips with known poses (Sec.~\ref{sec:results}).

\begin{proposition}[Exact repair]
\label{prop:repair}
If both behavioural classes are recovered, $\phi_E$ is invertible with respect to $\phi_T$, and the start pose is known, then $E_{\phi_T}\circ E_{\phi_E}^{-1}$ returns the training tensor for the same path, up to floating-point error.
\end{proposition}
\begin{proof}
Row-by-row inversion gives each chunk's displacement and heading change. Composing these from the start rebuilds the chunk-boundary poses, and invertibility means $\phi_T$ reads nothing else. Applying $E_{\phi_T}$ is then the training tensor by definition.
\end{proof}

\section{ActionFaultBench}
\label{sec:bench}
ActionFaultBench, our benchmark, consists of training--evaluation encoder pairs with known differences; it scores detection, naming, refusal and repair separately.

\topic{Categories.}
Scoring targets behaviour, not source edits. \emph{Clean} pairs behave the same through independent code and \emph{no-op} pairs change the source but not the behaviour; reporting a difference on either is a \emph{false alarm}. \emph{Identifiable} and \emph{non-invertible} pairs, together the \emph{differing pairs}, differ in family fields, with or without a way back from evaluation outputs; a differing pair is an \emph{exact recovery} when the named fields are exactly the true differences. \emph{Ambiguous} pairs restrict the paths so that some differences cannot be revealed, and naming a field outside the true differences is an error. \emph{Out-of-family} pairs must be refused, and \emph{noisy} pairs declare noisy poses.

\topic{Sealed runs.}
A \emph{sealed run} tests frozen code on a \emph{sealed set} of pairs its developers have not seen. Before each run we fixed the categories and targets ($\geq$95\% exact recovery, no wrong field named, $\geq$95\% refusal, floating-point repair) and froze the code by hash; a separate automated agent, without access to the method code or earlier sets, then wrote the pairs and labels in new \emph{interface families}, groups of pairs built around one encoder convention, from a description of our contract family (complete from the third run on). The sets thus test our family against itself, and refusal only on deliberate departures from it; no person outside the method team prepared or checked any set. Predictions were stored once before labels were opened. Runs 1--3 were development challenges, each followed by fixes, and stand as scored; the final run is the only test of witness-backed naming (details in \supp{L}). Pairs in one interface family are not independent, so we give counts, not binomial intervals. Faults injected into the encoders of NWM (Sec.~\ref{sec:models}), RAE-NWM~\cite{zhang2026rae}, Vid2World~\cite{huang2025vid2world} and the ViNT, GNM and NoMaD navigation policies~\cite{shah2023vint,shah2023gnm,sridhar2024nomad} were used in development, so results on them are optimistic. Baseline settings and the real-code pairs are in \supp{O}.

\section{How Well the Check Works}
\label{sec:results}
\begin{table}[t]
\centering\footnotesize
\setlength{\tabcolsep}{2.6pt}
\caption{\textbf{ActRosetta on the four sealed runs.} Counts are encoder pairs; runs 1--3 were development runs, each followed by fixes. Exact recovery is out of differing pairs, refusal out of out-of-family pairs; family mean and worst family are the mean and minimum share correct over interface families. Repair error is on pairs repaired from outputs. \icok/\icno: final-run target met/missed.}
\label{tab:sealed}
\begin{tabular}{lrrr>{\columncolor{Indigo!9}}rl}
\toprule
& \multicolumn{3}{c}{Development} & \textbf{Final} & Target\\
& Run 1 & Run 2 & Run 3 & \textbf{run} &\\
\midrule
Encoder pairs & 202 & 228 & 236 & \textbf{240} &\\
Interface families & 7 & 12 & 12 & \textbf{18} &\\
\rowcolor{black!5}\multicolumn{6}{l}{\emph{Right answers (more is better)}}\\
\quad Exact recovery & 70/76 & 70/86 & 79/90 & \textbf{73/91} &\\
\qquad as a rate & 92\% & 81\% & 88\% & \textbf{80\%} & \icno\ $\geq$95\%\\
\quad Correctly refused & 37/41 & 40/46 & 46/47 & \textbf{48/48} &\\
\qquad as a rate & 90\% & 87\% & 98\% & \textbf{100\%} & \icok\ $\geq$95\%\\
\quad All pairs correct & 185 & 200 & 224 & \textbf{221} &\\
\quad Family mean & .91 & .88 & .95 & \textbf{.92} &\\
\quad Worst family & .78 & .60 & .86 & \textbf{.62} &\\
\rowcolor{black!5}\multicolumn{6}{l}{\emph{Wrong field named (fewer is better)}}\\
\quad Clean, no-op pairs & 0/44 & 0/50 & 0/52 & \textbf{0/53} & \icok\ 0\\
\quad Ambiguous pairs & 0/29 & 5/32 & 0/33$^*$ & \textbf{0/34}$^*$ & \icok\ 0\\
\quad Differing pairs & 1/76 & 5/86 & 5/90 & \textbf{1/91} & \icno\ 0\\
\rowcolor{black!5}\multicolumn{6}{l}{\emph{Repair error from outputs (smaller is better)}}\\
\quad Median & 1.3e-14 & 3.8e-14 & 1.6e-14 & \textbf{9.8e-15} & \icok\ float\\
\quad Largest & 41.7 & 7.7e-6 & 2.9e-4 & \textbf{5.9e-8} &\\
\bottomrule
\end{tabular}

\smallskip{\footnotesize $^*$A ``no difference'' answer on a restricted domain is not counted as wrong; under run 2's stricter rule, run 3 gets 26/33.}
\end{table}
\topic{Final sealed run.}
The final run met the refusal target and named no wrong field on clean, no-op or ambiguous pairs, but recovered 73 of the 91 differing pairs exactly (80\%; Table~\ref{tab:sealed}). Of the 18 misses, 9 were refusals, 8 named too few fields or none, and 1 named a wrong field: a \emph{risk} of 1.2\% over the 84 pairs with any field named, at a \emph{coverage} of 88\% (80 of 91 differing pairs). Repair from outputs covered 70 pairs, 38 differing (Table~\ref{tab:base}) and 32 clean or no-op. Post hoc, the 29 differing pairs that combine two or three injected faults are harder to name exactly (19 of 29, against 54 of 62 single faults), but repair converts whole contracts: on all 16 of them repairable from outputs, including 7 where naming missed a field, it matched the training tensors to $4\times10^{-13}$.
\begin{table}[t]
\centering\footnotesize
\setlength{\tabcolsep}{2.3pt}
\caption{\textbf{Baselines on the final sealed set.} Counts are over its 91 differing and 53 clean and no-op pairs; ``names fields'' counts exact recoveries. Baselines ran untuned after the set was opened, at 280 transitions per pair; adapters fit paired outputs. Repair error: median over the 38 differing pairs ActRosetta could repair from outputs. $^\S$Post hoc, 976 transitions per pair.}
\label{tab:base}
\begin{tabular}{lrrrrc}
\toprule
Method & Notices & False & Names & Repair & Paired\\
& difference & alarms & fields & error & data?\\
\midrule
Random fuzzing & 90/91 & 0/53 & \icnone & \icnone & No\\
Family-free check$^\S$ & 91/91 & 0/53 & \icnone & \icnone & No\\
Affine adapter & -- & -- & \icnone & 1.9e-8 & Yes\\
MLP adapter & -- & -- & \icnone & 0.59 & Yes\\
\rowcolor{Indigo!9} \textbf{ActRosetta} & \textbf{91/91} & \textbf{0/53} & \textbf{73/91} & \textbf{1.2e-14} & No\\
\bottomrule
\end{tabular}
\end{table}
\topic{Baselines.}
Random fuzzing notices 90 of the 91 differing pairs with no false alarm, 84 already at 5 transitions (Table~\ref{tab:base}), so detection does not distinguish ActRosetta; fuzzing cannot say which fields differ or repair them. A fitted affine adapter needs paired outputs, and its largest error on the 38 pairs is 0.94, against 4e-13 for ActRosetta. ActRosetta costs more than fuzzing, a median of 400 transitions per pair.

\topic{Real code.}
On seven real training--evaluation pairs, with expected results written before running, witness-backed naming matched six and one partly. It named all three mismatches (frame, offset, yaw wrapping) in the executed NWM and MWM~\cite{yan2026mwm} rollouts and, by reading, in CompACT's~\cite{kim2026planning}, consistent with inherited code; named frame and offset but left yaw unresolved in World-in-World's released encoder; and found no mismatch in RAE-NWM, Vid2World and our patched World-in-World encoder. NWM's mismatches were named with witnesses of 5 or 10 transitions, but NWM informed development, so this shows impact, not generalization. From NWM's rollout outputs alone, the adapter reproduces the training tensors on 20 RECON clips to $5.7\times10^{-15}$.

\begin{table}[t]
\centering\footnotesize
\setlength{\tabcolsep}{2.2pt}
\caption{\textbf{Coverage audits of repositories not examined before each audit.} Locked: list fixed before any encoder code was read. $^\ast$Verdicts on real columns, out of the new repositories deriving actions from poses (4, 4 and 20 of 8, 8 and 24). Found by: 1, family-free check; H, also by reading. Evaluation: does a mismatch reach a model evaluation; dep.: deployment only.}
\label{tab:cov}
\begin{tabular}{@{}lcccp{.25\linewidth}p{.2\linewidth}@{}}
\toprule
Audit & New, seen & Locked & Verdicts$^\ast$ & Mismatches (found by) & Evaluation\\
\midrule
1 & 8, 0 & No%
& 3/4 & LogoNav (1, H)& None (dep.)\\
2 & 8, 0 & Yes & 3/4 & UMI timing (1, H)& None (dep.)\\
3 & 24, 16 & Yes & 8/20 & LCVN, ViPlanner, OpenDWM (1) & LCVN unmeasured\\
\midrule
All & 40 new & & 14/28 & 5, all by 1 & One candidate\\
\bottomrule
\end{tabular}
\end{table}
\topic{Coverage of new repositories.}\phantomsection\label{sec:coverage}
To test ActRosetta beyond our family, we audited 40 unexamined public repositories, 28 of which derive actions from poses (Table~\ref{tab:cov}). The family-free check, run on 26 of these, found five mismatches, and our non-exhaustive reading found none it missed. The second level reached a \emph{verdict}, any result other than refusal on the encoder's real output columns, on 14 of the 28, mostly weak ``no difference'' tests (\supp{G}), and refused all five mismatches. Only LCVN's mismatch may affect evaluation: rollouts imagined by its world model, where its actor-critic policy learns, receive actions one row ahead of the world model's training rows, which by our reading reaches its evaluation script. On its released data (3,602 trajectories), these inputs differ from training rows by a median relative difference of 0.504 [0.501, 0.507]; with no public checkpoint, the effect on predictions is unmeasured. LCVN is 1 of the third audit's 24 new repositories (95\% Clopper--Pearson interval 0.1--21.1\%).%

\section{Conclusion}
Rollout evaluation assumes that actions are encoded as in training. NWM's released rollout code breaks it through three interacting mismatches; repairing them, with unchanged weights, lowers NWM's mean DreamSim error on RECON by 42\% and changes how its checkpoints (exploratory) and alternative models compare. ActRosetta turns this audit into a check others can run on their own encoders. Before reporting rollout error, compare training and evaluation action encoders on paths longer than one chunk; if they differ, use the training code or identify every differing field before patching. Code and ActionFaultBench will be released.%

\topic{Limitations.}
Model evidence is concentrated on NWM and RECON. Repair changes error little where predictions respond weakly to actions (NWM on SCAND and Go Stanford), fault tests on alternative models use only 4\,s clips, and repair did not measurably improve decisions in Habitat, so we claim nothing about control. ActRosetta names and repairs only within a finite planar family. Exact recovery missed our 95\% target in every sealed run (73 of 91 in the final one), and in 40 new repositories naming refused all five mismatches that the family-free check found, with a second-level verdict on 14 of 28. Automated agents wrote the sealed sets from our family's description, so they test the family against itself. Our diagnosis rests on NWM's code at the audited commit; later versions may differ.

\section*{LLM Usage Statement}
OpenAI Codex and ChatGPT assisted with research planning, experiment and analysis scripts, literature checks and earlier drafts. Anthropic Claude Code implemented ActRosetta and ActionFaultBench, ran experiments and analyses, drew figures, wrote the sealed test sets through separate agent instances, and drafted, restructured and edited this manuscript. GPT and Claude produced critical reviews of earlier drafts that shaped this revision. Numerical claims are drawn from retained outputs, and figures plot measured data or diagrams, not generated experimental imagery. OpenAI image generation produced only the decorative robot icons in Figure~\ref{fig:method}. The authors remain responsible for all claims, references, code and the final submission.
{\small\bibliographystyle{ieeenat_fullname}\bibliography{references}}

\clearpage
\appendix
\renewcommand{\thefigure}{S\arabic{figure}}\renewcommand{\theHfigure}{S\arabic{figure}}
\renewcommand{\thetable}{S\arabic{table}}\renewcommand{\theHtable}{S\arabic{table}}
\renewcommand{\thealgorithm}{S\arabic{algorithm}}\renewcommand{\theHalgorithm}{S\arabic{algorithm}}
\setcounter{figure}{0}\setcounter{table}{0}\setcounter{algorithm}{0}
\twocolumn[{\ppappendixbanner}]
These appendices give the settings, intervals and additional results behind the main paper. Section, figure and table numbers without an ``S'' or a letter refer to the main paper. Section~3 of the main paper, on how action encoding changes NWM's evaluation, is supported by Secs.~\ref{app:faults}--\ref{app:fid}, \ref{app:ctl}, \ref{app:grid}, \ref{app:moved} and \ref{app:plan}--\ref{app:faultx}. Sections~4 and~5, on contracts and ActRosetta, are supported by Secs.~\ref{app:family} and~\ref{app:probe}. Sections~6 and~7, on ActionFaultBench and its results, are supported by Secs.~\ref{app:base}, \ref{app:cov}, \ref{app:sens}, \ref{app:history} and~\ref{app:proto}. As in the main paper, \emph{released actions} are built by NWM's released rollout code and \emph{repaired actions} by its training code from the same poses; for an alternative model, \emph{correct actions} are those built by the model's own encoder, and for NWM they are the repaired actions. Brackets give 95\% confidence intervals from a bootstrap that resamples whole trajectories (written ``trajectory-level'' or ``trajectory-cluster'' below) unless a section states otherwise; scene-cluster and day-level intervals resample simulator scenes and recording days instead.

\section{Baselines, pose noise and the family-free check}
\label{app:base}
\begin{figure*}[t]
\centering\includegraphics[width=\textwidth]{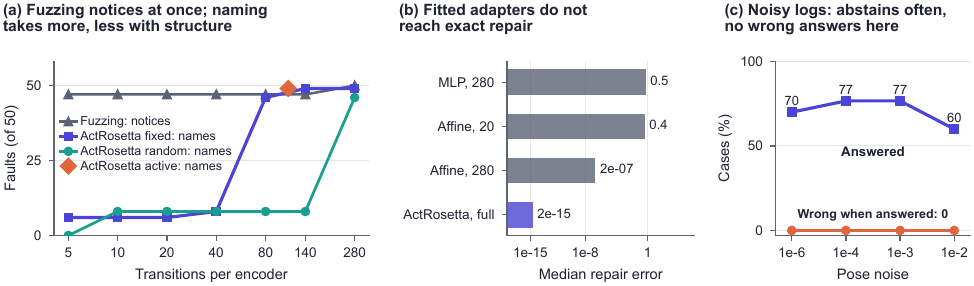}
\caption{\textbf{Detection, naming, repair and pose noise on the development interfaces} (50 injected faults in the training encoders of NWM, RAE-NWM, Vid2World and ViNT; seen during development, so optimistic). (a) Detection finds that two encoders differ; naming identifies the injected fault exactly. Random fuzzing detects within a few transitions, while ActRosetta's naming needs more transitions per encoder, fewest with active probing (Sec.~\ref{app:probe}): 49 of 50 named at a median of 110 transitions, with no wrong field. (b) Median action-repair error (log scale) at chunk size 4. Labels give each adapter's training transitions; ActRosetta uses its full probe set. Adapters are scored by median and ActRosetta by maximum absolute error per case (see text). (c) Share of 120 noisy cases (two interfaces, three resampled probe sets each) that ActRosetta answers, by pose-noise level, the standard deviation of Gaussian noise added to logged positions and headings; no answer was wrong. Removing the noisy-mode stability gate (Sec.~\ref{app:probe}) also produced no wrong answer here.}
\label{fig:base}
\end{figure*}

This section compares ActRosetta with simple baselines on the public interfaces used during development, where the injected faults are known, and shows how it behaves when logged poses are noisy. These interfaces informed the method, so all numbers here are optimistic; the held-out comparison is Table~3. All tests call deterministic encoders on synthetic paths, so no seeds or diffusion steps apply.

\topic{Naming and repair.}
Figure~\ref{fig:base}a separates detection from naming on 50 injected faults in four interfaces, the training encoders of NWM, RAE-NWM, Vid2World and ViNT. Random fuzzing detects most faults within a few transitions but names none. With counterexample-guided probing (Sec.~\ref{app:probe}), ActRosetta names 49 of the 50 faults at a median of 110 transitions, with no wrong field; on random paths it names 46 at 280 transitions, and its fixed probes name 49 at 140. In Figure~\ref{fig:base}b, each fitted adapter is scored by its median absolute error per action element, while ActRosetta is scored by its maximum absolute error over the test paths, the stricter of the two. The plotted values are medians over cases, so the two scores aggregate differently within a case.

\topic{Pose noise.}
Figure~\ref{fig:base}c tests the noisy mode (Sec.~\ref{app:probe}) on 120 cases from two interfaces (NWM's and RAE-NWM's training encoders, at noise levels $10^{-6}$, $10^{-4}$, $10^{-3}$ and $10^{-2}$), with three resampled probe sets per case rather than the five used in the sealed runs. ActRosetta answers 60--77\% of the cases, depending on the noise level, and none of its answers is wrong. Section~\ref{app:probe} defines the stability gate. On the noisy pairs of the four sealed runs, ActRosetta was correct on 12 of 12, 13 of 14, 14 of 14 and 13 of 14; one of the two misses, in run~2, was a wrong answer.

\topic{The family-free check on the final sealed set (post hoc).}
To see what detection alone achieves on held-out pairs, we ran ActRosetta's family-free check by itself on all 240 pairs of the final sealed set, after the set had been opened. It is the differential test of the refusal branch of Algorithm~1 (line~2): both encoders are called on a fixed set of 24 paths (the 18 default probes, 4 random hold-out paths and 2 long paths of 48 transitions each, at the pair's chunk sizes; this set is separate from the range and hold-out paths that recovery uses in Sec.~5), and any relative output difference above $10^{-6}$, change of output shape or exception counts as a difference. Where a pair restricts the allowed paths, only those paths are used. The check flagged none of the 53 clean and no-op pairs, all 91 differing pairs (as did ActRosetta with naming; Table~3) and all 48 out-of-family pairs. Ambiguous pairs have no detection label; 10 of the 34 differ on their restricted paths. All 14 noisy pairs differ, as expected when logged poses are noisy, so the check does not apply to them. It used a median of 976 transitions per pair, counted as ActRosetta's recovery counts them, against 400 for ActRosetta and 280 for random fuzzing. Because the set had been opened, this is not a held-out number; the frozen code's hashes were unchanged before and after the run. On this set, the family-free check thus adds no detection beyond ActRosetta's own.

\section{Injected faults in NWM, and why the benefit of repair varies}
\label{app:faults}
\begin{figure*}[t]
\centering\includegraphics[width=\textwidth]{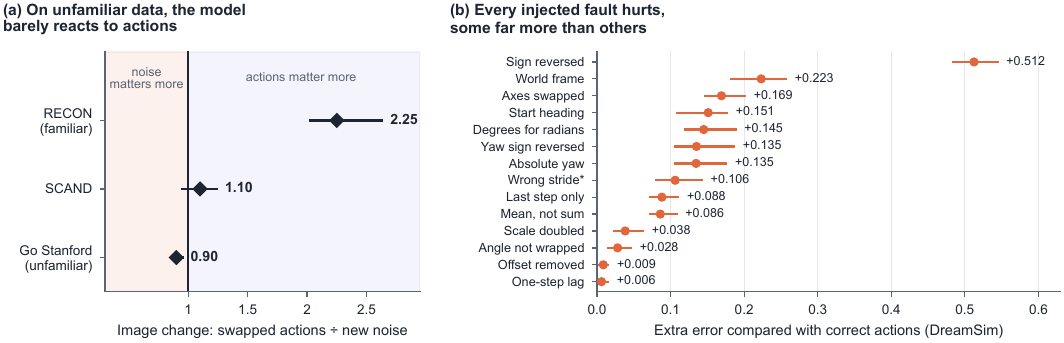}
\caption{\textbf{NWM's action responsiveness and the damage from injected faults.} (a) Action-responsiveness ratio $R=\bar E/\bar F$: the image change caused by substituting another trajectory's actions of matched path length, divided by the change caused by a new noise seed (defined in Sec.~\ref{app:faults}). Above~1, actions move predictions more than noise does. 50 clips per dataset (14, 12 and 40 trajectories for RECON, SCAND and Go Stanford), seeds 0--2, 50 diffusion steps. (b) Increase in mean DreamSim error over 1--16\,s caused by each injected fault, relative to correct actions (for NWM, the repaired actions; mean error 0.135): 50 main-test RECON clips, 15 trajectories, seed~0, 50 diffusion steps. Bars: 95\% trajectory-level confidence intervals. $^*$The stride fault needs 32\,s of poses, which 25 of the clips lack; their missing poses repeat the last pose, and the text reports the 25 complete clips.}
\label{fig:response}
\end{figure*}

This section gives the details of the injected-fault test on NWM and of the action-responsiveness ratio, both used in Sec.~3 (``Injected faults also hurt the alternative models'' and ``Where repair matters'').

\topic{Injected faults.}
We injected 14 faults, each at one setting, into NWM's repaired actions on 50 main-test clips (15 trajectories, 1\,fps, 16\,s, 50 diffusion steps, seed~0). Every fault raised mean DreamSim error, with every 95\% interval above zero, from +0.0062 for a one-step lag to +0.5125 for a reversed sign (Figure~\ref{fig:response}b). The stride fault, under which each chunk reads twice as many raw time steps, needs 32\,s of poses; on the 25 clips that have them, which span only four trajectories, it raises error by +0.0835 [+0.0583, +0.1238]. Each fault was run at one parameter setting, so the ordering in Figure~\ref{fig:response}b need not hold at other magnitudes.

\topic{Repair of the injected faults.}
For all 14 faults on all 50 clips, ActRosetta's repaired actions equal the correct actions as float32 values: 11 faults were repaired by the adapter from evaluation outputs, and 3 (lag, stride and last-step aggregation, which are not invertible) from poses. Because NWM's generation is deterministic, equal actions give the same rollout. We confirmed this by generating 3 clips for each of 3 repaired faults, which showed zero pixel difference.

\topic{Correcting one mismatch at a time at 4\,fps.}
Section~3 compares frame-only correction with full repair at 1\,fps. At 4\,fps (the same 50 clips, 64 frames over 16\,s, 50 diffusion steps, seed~0), correcting only the reference frame leaves DreamSim error 0.058 [0.036, 0.093] above full repair. Correcting only yaw changes little, even though the released yaw mismatch produces jumps of a full turn in 63 of the 150 main-test clips; Sec.~\ref{app:mech} discusses why.

\topic{Why the benefit of repair varies across datasets.}
Repair helps far less outside RECON. On Go Stanford, which NWM was not trained on, mean DreamSim error moves from 0.505 to 0.500, a reduction of 0.005 [0.001, 0.010] (150 clips, 79 trajectories, 50 diffusion steps, seed~0), although the path that the released actions command ends a mean 3.76\,m [3.36, 4.17] from the recorded endpoint at 16\,s. SCAND, which is in NWM's training data, is similar: 0.319 to 0.313, a reduction of 0.006 [0.001, 0.014] (150 clips, 13 trajectories, 25 diffusion steps, seed~0).

The action-responsiveness ratio suggests why (Figure~\ref{fig:response}a). For each clip, $E$ is the DreamSim distance between the predictions made from the clip's own actions and from another trajectory's actions of matched path length, at the same noise seed, and $F$ is the distance between the predictions made from the clip's own actions under two different seeds; $R=\bar E/\bar F$, the ratio of their means over clips and seeds. We fixed its reading before running: a lower confidence bound above 1.5 means that images change with actions beyond noise, an upper bound below 1.25 means that the action effect is indistinguishable from or below noise, and anything else is inconclusive. On RECON, $R$ is 2.25 [2.03, 2.63]; on SCAND, 1.10 [0.95, 1.24]; and on Go Stanford, 0.90 [0.86, 0.95]. Where images barely respond to actions, correcting the actions has little room to help, so a small effect of repair is no evidence that a mismatch is harmless.

\topic{Responsiveness by checkpoint.}
To test whether responsiveness explains why the Ego4D-trained checkpoint gains least from repair on the new cohort (Sec.~3), we computed $R$ for all five checkpoints on the same 50 RECON clips (14 trajectories, seeds 0--2, 50 diffusion steps, 95\% trajectory intervals). It is 2.25 [2.03, 2.63] for XL, 2.36 [2.12, 2.72] for L, 1.81 [1.65, 2.07] for B, 1.37 [1.24, 1.56] for S and 1.40 [1.28, 1.60] for XL trained with Ego4D. Their mean DreamSim reductions from repair on the new cohort are 0.076, 0.100, 0.093, 0.080 and 0.0076. S and the Ego4D checkpoint respond alike to actions but gain very differently, so responsiveness does not explain the Ego4D checkpoint's small gain. What separates these two checkpoints from the others is that a new seed changes their frames more ($\bar F$ of 0.17--0.18, against 0.10--0.13), not that substituted actions change them less. XL's ratio was computed on another machine with the same GPU model. The other four checkpoints share one public training dataset class, so one contract check covers them. No public code builds the Ego4D checkpoint's action-labelled training data, and its checkpoint file keeps no training settings, so its contract cannot be checked.

\section{Vid2World: injected faults, repair and responsiveness}
\label{app:v2w}
\begin{figure}[t]
\centering\includegraphics[width=\linewidth]{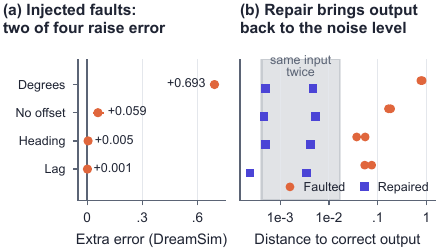}
\caption{\textbf{Injected faults and their repair in Vid2World on RECON.} (a) Increase in mean DreamSim error over 16 predicted frames (4\,s at 4\,fps) relative to correct actions, built by Vid2World's own dataset code: 30 clips, 19 trajectories, one seed, 50 diffusion steps; bars are 95\% trajectory-level confidence intervals. (b) DreamSim distance from the predictions made with correct actions, for faulted and repaired actions (two clips per fault; log scale). Vid2World is not bit-deterministic, so the shaded band shows the distance between two runs with identical input and seed. Repaired predictions differ from the correct ones about as much as such a repeated run does, and far less than faulted ones.}
\label{fig:second}
\end{figure}
Vid2World~\cite{huang2025vid2world} uses the same action normalization as NWM but is a separately trained video model, so it tests whether NWM's encoding choices matter in a second predictor (Sec.~3, ``Where repair matters''). We use its RECON checkpoint, with its own dataset code building the correct actions, on 30 RECON clips from 19 trajectories: one seed (equal to the clip index), 50 diffusion steps with DDIM sampling~\cite{song2021denoising} and classifier-free guidance 2.5, and 16 predicted frames at 4\,fps (4\,s). The first 10 clips were extended by 20 clips from new trajectories, chosen by a rule fixed before generation but not at random; the extension took 13.70 GPU hours on one RTX 3090 Ti, in one process. Intervals are 95\% trajectory-cluster bootstrap intervals, without correction for multiple comparisons.

\topic{Injected faults.}
Removing the normalization offset raised DreamSim error by 0.059 [0.039, 0.083], from 0.085 with correct actions, and it raised error on all 30 clips. Supplying yaw in degrees instead of radians raised error by 0.693 [0.675, 0.708]; in the clips we inspected, the predictions became texture rather than scenes, so this measures a broken model, not a realistic degradation. Two faults had small effects: a reference frame fixed at the start heading (+0.005 [$-$0.002, +0.012]) and a one-step lag (+0.001 [$-$0.003, +0.005]). Their LPIPS intervals exclude zero but their DreamSim intervals do not, and none of the eight intervals (four faults, two metrics) is corrected for multiple comparisons, so we read these two effects as weak evidence.

\topic{Repair.}
ActRosetta diagnosed all four faults correctly (offset, frame, yaw units and lag). Its repaired tensors match the correct ones to at most $4.5\times10^{-8}$; only the lag, which is not invertible and was repaired from poses, is bit-identical on every clip. Because Vid2World is not bit-deterministic, two runs with identical input and seed give slightly different frames. After finding this, we compared regenerated frames with a repeated run of the correct input, on 2 clips per fault (Figure~\ref{fig:second}b). On the two clips, repaired predictions were at DreamSim 0.0002--0.0005 and 0.004--0.005 from the correct ones, against 0.0004 and 0.017 for a repeated run with the same input, and 0.037--0.82 for faulted predictions. Over 10 clips, repeated runs with the same input changed the error to ground truth by a mean of 0.0007 DreamSim and 0.0009 LPIPS, and all 10 showed some non-determinism.

\topic{Action responsiveness.}
With $E$, $F$ and $R$ defined as in Sec.~\ref{app:faults}, we measured Vid2World's responsiveness on 12 of the 30 clips, chosen by a seeded rule (9 trajectories), with seeds 0--2 at the settings above; donors were path-length-matched clips from other trajectories. Substituting a donor's actions changed the frames 2.54 [2.16, 3.02] times as much as a new seed did by DreamSim ($\bar E=0.1229$, $\bar F=0.0484$; $\bar E-\bar F=+0.0745$ [+0.0564, +0.0958]; $E>F$ on all 12 clips), and 1.84 [1.70, 1.98] times as much by LPIPS (95\% trajectory-cluster bootstrap). By the reading fixed in advance, Vid2World's images change with actions beyond noise. The seed-0 predictions with correct actions were reused from a run on another machine; $\bar F$ is 0.0490 with them and 0.0473 without. NWM's ratio on RECON, 2.25 [2.03, 2.63] (Sec.~\ref{app:faults}), comes from other clips at 1\,fps over 16\,s, so the two ratios are compared only descriptively. Vid2World's ratio rests on 12 clips from 9 trajectories, a 4\,s horizon, one checkpoint and RECON only.

\section{DrivingWorld: injected faults in a tokenized driving model}
\label{app:dw}
DrivingWorld~\cite{hu2026drivingworld} is a driving world model, not a robot-navigation one, and it takes actions as discrete tokens of metres and of yaw in degrees. We included it to test whether injected faults hurt a model whose action format differs from NWM's (Sec.~3, ``Where repair matters''). It generates tokens autoregressively with top-$k$ sampling ($k=30$), so diffusion steps do not apply.

\topic{Injected faults.}
We injected four faults into its 12 moving demonstration clips, which its authors chose, and generated 40 frames (8\,s at 5\,Hz) with one seed; intervals are 95\% bootstrap intervals over clips. We fixed this design after a two-clip smoke test in which one fault had already lowered error. ActRosetta named all four faults on the continuous actions before tokenization, but it refuses the token stage as outside its family and abstains from repair at the model's one chunk size, so repair used DrivingWorld's own encoder on the poses. Here, injected faults did \emph{not} clearly hurt. Supplying yaw in radians instead of degrees, a one-step lag and a world reference frame changed DreamSim error by $-$0.000 [$-$0.005, +0.004], +0.002 [$-$0.001, +0.005] and $-$0.001 [$-$0.003, +0.003], and they changed the frames no more than a new seed does. Doubling the forward distance \emph{lowered} error by 0.037 [0.020, 0.059], on 11 of 12 clips. On 2 clips, the model's own forward prediction from real context frames was 0.54--0.86 of the recorded value. Either the model under-advances, or the demonstration actions differ in scale from its training actions; with no public training script, we could not check which. This run therefore does not show that injected faults degrade evaluation in this format, and it shows that lower image error cannot replace an interface check.

\topic{Follow-up on responsiveness and scale.}
After seeing that result, we fixed a follow-up design in advance on the same 12 clips and 40 frames, with one seed for the substituted actions, one seed pair for noise, and a bootstrap over clips (with 12 clips, the intervals are approximate). Another clip's length-matched actions changed the frames by DreamSim 0.0542, against 0.0545 for a new seed: $R=0.995$ [0.768, 1.224]. By the reading fixed in advance (Sec.~\ref{app:faults}), the action effect is indistinguishable from noise; LPIPS gives 1.137 [0.862, 1.406], which is inconclusive. Error fell steadily as the commanded distance grew: 0.149, 0.133, 0.107 and 0.096 at 0.5, 1, 1.5 and 2 times the given distance. Relative to the given actions, half the distance raised error by 0.016 [0.008, 0.026], and double the distance lowered it by 0.037 [0.020, 0.059]. Progress along the recorded clip, fitted from DreamSim embeddings, rose from 0.44 [0.37, 0.52] with the given actions to 0.89 [0.74, 1.00] at double distance, while sharpness fell slightly. Lower error thus came with frames that keep pace with the recorded clip, not with sharper frames. This fits demonstration actions that differ in scale from training, or a model that under-advances; without a training script, we cannot tell which. The tokens and the doubled-distance result reproduced the earlier run on all 12 clips, with identical frames on 11.

\section{Anchoring to NWM's published numbers}
\label{app:anchor}
\begin{table*}[t]
\centering\footnotesize
\setlength{\tabcolsep}{4pt}
\caption{\textbf{RECON rollout error with NWM's own evaluation script, against NWM's published values.} 150 main-test clips (20 trajectories), 1\,fps, seed~0, one sample, 250 diffusion steps. Published: mean $\pm$ std over five samples for the checkpoint of NWM's paper~\cite{bar2025navigation}, as its authors posted them~\cite{nwm2025issue21}; our values use the released checkpoint, which was retrained after face blurring. Released: released actions with \texttt{torch.compile} on, as the released script runs. Repaired: repaired actions with compile off (compile on was run for only 74 clips, on which it moves repaired error by at most 0.002; see text). Released and repaired values are the script's own means, which weight batches equally (see text). Gap: released minus repaired LPIPS, both with compile off, as plain clip means with 95\% trajectory-level intervals. FID has no intervals, because our own feature computation differs from the script's FID by about 1.5.}
\label{tab:anchor}
\begin{tabular}{lrrrrrrr}
\toprule
& \multicolumn{4}{c}{LPIPS} & \multicolumn{3}{c}{FID}\\
\cmidrule(lr){2-5}\cmidrule(lr){6-8}
Horizon & Published & Released & Repaired & Gap & Published & Released & Repaired\\
\midrule
1\,s & 0.258$\pm$0.001 & 0.273 & 0.273 & 0.000 & 43.05$\pm$0.59 & 48.07 & 47.86\\
2\,s & 0.295$\pm$0.004 & 0.315 & 0.302 & +0.009 [0.004, 0.014] & 46.50$\pm$0.32 & 50.16 & 49.99\\
4\,s & 0.370$\pm$0.002 & 0.369 & 0.328 & +0.041 [0.029, 0.057] & 51.25$\pm$0.83 & 55.95 & 52.72\\
8\,s & 0.470$\pm$0.001 & 0.458 & 0.377 & +0.079 [0.058, 0.102] & 65.55$\pm$1.12 & 78.11 & 54.34\\
16\,s & 0.574$\pm$0.003 & 0.550 & 0.430 & +0.121 [0.097, 0.143] & 93.13$\pm$2.00 & 88.66 & 64.27\\
\bottomrule
\end{tabular}
\end{table*}
Section~3 states that, with NWM's own evaluation script, our released-action LPIPS is close to NWM's published values. This section gives the full comparison (Table~\ref{tab:anchor}), the tests we set for it in advance, and two details of the released code that it exposed. NWM publishes LPIPS and FID for RECON rollouts, not DreamSim, so the comparison uses those two metrics.

\topic{How we ran NWM's script.}
NWM's released weights were retrained after face blurring, as its maintainer states~\cite{nwm2025issue26}, so they are not the paper's checkpoint. NWM's inference script does not run as released, because an import is commented out. We therefore ran the unedited file through a small runner that adds only that import, a fix for batch size one, a per-clip seed and the data path. The repaired arm changes only the action deltas.

\topic{Tests set in advance.}
The first test asked whether our released-action values lie within three published standard deviations; they do not, except for LPIPS at 4\,s and FID at 16\,s, so this is not a reproduction within the paper's sampling noise. The second test allowed for the retrained checkpoint, with a margin of 10\% for LPIPS and 20\% for FID; every horizon passes, the closest being FID at 8\,s, at +19.2\%. Released and repaired actions give identical results at 1\,s, where the mismatches cannot act. On the time-shift evaluation, which predicts one frame a fixed time ahead in a single call and so cannot be reached by the mismatches, our results with the released checkpoint (500 clips) match numbers that another user posted publicly for it~\cite{nwm2025issue26}, within limits we set in advance; for instance, LPIPS at 4\,s is 0.309 against 0.310. At 8 and 16\,s, our released-action LPIPS is within 5\% of the published values, and repaired-action LPIPS is 20\% and 25\% below them (Sec.~3). We read this as an anchor with an offset that the checkpoint difference could explain but that we cannot test: it is consistent with, but does not prove, the published results having used the released encoding. We ran neither the 4\,fps setting nor the other four published samples.

\topic{Two details of the released code.}
First, the released script calls \texttt{torch.compile}, which changes rollouts. On the same seed, released 16\,s error is lower with compile on, by 0.0072 [0.0024, 0.0128] LPIPS and 0.0136 [0.0037, 0.0261] DreamSim (150 clips); other horizons move by less than 0.001. Repaired error moves by at most 0.002 at any horizon (74 clips from 17 trajectories; the run stopped at our compute cap). The main paper's 0.240 is a compile-off number. On the 22 clip--arm pairs we checked, frames generated with compile off matched across two machines to within one intensity level. Second, the evaluation script averages batch means: with 150 clips in batches of 64, its last batch of 22 clips counts as much as a full batch. This moves means by up to 0.011 and makes them depend on the order in which files are listed, so we report plain clip means everywhere else.

\section{Action fidelity}
\label{app:fid}
Lower image error need not mean that predictions follow the commanded motion. Section~3 reports three tests of whether they do; this section gives their settings and full results. Here the \emph{true path} is the recorded path, and the \emph{misread path} is the path that the released actions describe when they are decoded with the training contract, that is, the motion a model that reads its inputs as in training would be asked to imagine. All three tests were locked before any result existed. The inverse dynamics model failed its locked forward-motion check, so its results are exploratory; the other two tests do not depend on it. Intervals are 95\% bootstrap intervals over trajectories (10,000 resamples).

\topic{Counterfactual retrieval.}
For the first 50 main-test clips (15 trajectories), we chose four other clips from poses alone: from other trajectories, with 16\,s endpoints at least 2\,m from the true one. Each clip's own actions and the four other clips' actions generated five 16\,s futures per arm (500 rollouts in all; 50 diffusion steps, 1\,fps, seed equal to the clip index), and we asked whether the future generated from the clip's own actions is the one closest to the recorded 16\,s frame (chance 0.20). By DreamSim, it was closest for 15 of 50 clips with released actions and 40 of 50 with repaired actions; averaged over trajectories, the accuracies are 0.50 [0.31, 0.69] and 0.86 [0.71, 0.97], a gain of +0.36 [+0.10, +0.59]. By LPIPS, the counts are 10 and 38 of 50 and the accuracies 0.32 [0.14, 0.52] and 0.75 [0.58, 0.90], a gain of +0.43 [+0.08, +0.73]. The DreamSim margin of the mean other clip over the clip's own actions was 0.061 [0.005, 0.118] with released actions and 0.162 [0.132, 0.191] with repaired actions. With repaired actions, the commanded motion is therefore visible in the 16\,s image.

\topic{Inverse dynamics.}
An inverse dynamics model (a ResNet-34~\cite{he2016deep} that predicts the motion between two frames) was trained on 1,470 RECON trajectories from 6 recording dates and validated on all trajectories of a seventh date. Its training data share no trajectory with any test list and no date with the main test. Before use, it had to pass three checks on real frames: yaw, forward motion and a path test. It passed the yaw and path checks, but in all three allowed attempts it failed the forward-motion check, which required the 95\% interval of the slope of estimated on true forward motion to lie inside [0.8, 1.2] (best 0.693 [0.582, 0.804]). Forward motion over one chunk barely varies (mean 0.996\,m, SD 0.128\,m), which may make that check unreachable; we noticed this only after the failures.

We applied the model to the main-test rollouts (150 clips, 20 trajectories, 250 diffusion steps, seed~0). Distances below are mean position errors over 1--16\,s between the estimated path and a reference path, without alignment. On real frames of the test clips, the estimated path is 1.49\,m [1.13, 1.90] from the true path. With released actions, the estimated path is 3.81\,m [3.13, 4.51] from the true path and 3.62\,m from the misread one; with repaired actions, it is 2.42\,m [1.77, 3.12] and 4.06\,m. Repaired predictions are therefore 1.39\,m [0.89, 1.87] closer to the true path than released ones (18 of 20 trajectories). Measured as distance to the true path minus distance to the misread path, released predictions lean towards the misread path by only +0.19\,m [$-$0.29, +0.79] (9 of 20 trajectories nearer the misread path), while repaired predictions lean towards the true path by 1.64\,m [1.00, 2.22] (1 of 20 nearer the misread path). Released predictions are nearer the true path at 4\,s and nearer the misread one only by 16\,s, so they leave the true path without clearly following the misread one. Regressing estimated on commanded yaw per chunk gives a slope of 0.741 [0.682, 0.810] with released actions and 0.965 [0.944, 0.990] with repaired actions. The model reads forward motion 0.35\,m [0.21, 0.52] less accurately on generated than on real frames, and the image decoder's round trip alone adds only 0.008\,m, so the instrument's error on generated frames cannot be separated from the world model's.

\topic{Geometric relative pose.}
As a check that does not depend on a learned model of motion, DISK features~\cite{tyszkiewicz2020disk} matched with LightGlue~\cite{lindenberger2023lightglue} recover the relative pose between frame pairs, with a fisheye focal length fitted on training trajectories only. On real frames, they succeed on 83\% of pairs, with a median yaw error of 2.4$^\circ$. On generated frames of every fifth main-test clip (30 clips, 11 trajectories), they succeed on only 12\% of released and 17\% of repaired pairs. On chunks where the true and misread headings differ by at least 20$^\circ$, released frames point 28.8$^\circ$ [8.6, 49.1] nearer the misread heading (7 of 9 trajectories), and repaired frames 30.1$^\circ$ [7.9, 55.5] nearer the true one (8 of 10). Coverage is low and may favour clearer frames, so we read this only as thin support. Together, the three tests support that repaired predictions follow the true path; that released predictions follow the misread path is supported only by this geometric check and, at 16\,s, by the inverse dynamics model.

\section{Coverage audit of 40 repositories}
\label{app:cov}
Section~7 audits 40 public repositories that had not been examined before each audit (Table~4); Table~\ref{tab:repos} lists them with their pinned commits. This section defines the terms used there, describes the two levels of the check and the three audits, and explains why most verdicts are weak tests. It ends with LCVN, the one mismatch that may reach an evaluation.

\topic{Terms.}
The \emph{real columns} are an encoder's output columns exactly as its code produces them. A \emph{verdict} is any comparison result on the real columns other than refusal: the two encoders are the same, show no observed difference, or differ. A mismatch is \emph{known} when either level of the check or our reading found it; there is no ground truth for mismatches that all three missed.

\topic{Two levels on real repositories.}
Of the 40 repositories, 28 derive actions from poses. ActRosetta's family-free check, its first level, ran on 26 of them (all 8 from the first two audits and 18 of the 20 from the third); Senna and MAD use one function for both sides, so there was nothing to compare. On all five known mismatches (LogoNav's Frodobot script, UMI's timing pair, LCVN, ViPlanner and OpenDWM), the frozen code refused the real columns, and its refusal note recorded that the two encoders differ, which is the family-free check's answer; a separate audit script, applying the same test, recorded the path on which the encoders differ. The check also flagged OccWorld, but only under a convention we assumed, so that case is not counted. The second level reached a verdict on the real columns for 14 of the 28. We began reporting the two levels separately only after the sealed tests.

\topic{First two audits.}
On eight public repositories not used before, ActRosetta reached a verdict on three. On eight more, whose list was fixed before any encoder code was read, it again reached a verdict on three (TransFuser, DiffusionDrive and PLUTO), all ``no observed difference''. We also tried a wider contract family, designed on the first audit's repositories. It raised the first audit's verdicts from three to six, of which only four do not rest on our own assumptions or placeholder values, but it added no verdict in the second audit; all reported results therefore use the family of Sec.~4 with witness-backed naming. LogoNav's mismatch, goal yaw converted to radians twice, is in a deployment script, not a model evaluation. UMI's timing difference between training and deployment appears to be a deliberate design choice, so a detected difference is not necessarily a bug.

\topic{Third audit.}
Before any encoder code was read, the list was fixed: the 16 repositories of the first two audits, followed by 24 new ones taken in salted-hash order from a pool of 48 that an automated agent had built, partly from memory. No listed repository was replaced. Commits are pinned, expected results were hashed before each run, and nothing was installed or imported; some encoders were transcribed by hand rather than executed. Of the 24 new repositories, 21 have complete code and 20 derive actions from poses. ActRosetta reached a verdict on the real columns for 8 of these 20, and for 11 if we may drop columns, and it named no field on real columns. The family-free check found 3 mismatches: one may affect evaluation (LCVN, provisional), one affects deployment only and looks intended (ViPlanner), and one is in code that never runs (OpenDWM). On the real columns of all five mismatches, including the two in previously examined repositories, ActRosetta refused. Among the 16 previously examined repositories, LogoNav and UMI showed deployment-only mismatches. LCVN comes from the same lab as UniWM, one of the first audit's repositories, and its actor--critic code has an NWM path, so this one candidate does not show that such mismatches reach beyond NWM's neighbourhood. Several repositories share lineage, so they are not independent, and three-dimensional motion was not modelled.
\begin{table*}[t]
\centering\footnotesize
\caption{\textbf{The 40 audited repositories and their pinned commits} (GitHub \texttt{owner/name}, first 7 characters of the commit). The first two audits' 16 repositories are also the 16 previously examined repositories of the third audit; the second audit additionally read \texttt{motional/nuplan-devkit} (e924167) as a helper for DiffusionDrive's evaluation geometry.}
\label{tab:repos}
\begin{tabular}{@{}>{\raggedright\arraybackslash}p{.1\linewidth}>{\raggedright\arraybackslash}p{.86\linewidth}@{}}
\toprule
Audit & Repositories\\
\midrule
First (8) & \texttt{F1y1113/UniWM} 747d7a8; \texttt{YvanYin/DrivingWorld} faceb98; \texttt{NHirose/Learning-to-Drive-Anywhere-with-MBRA (LogoNav)} c39384a; \texttt{ai4ce/CityWalker} ab0ef60; \texttt{NVlabs/X-MOBILITY} 5837aea; \texttt{gaoyuezhou/dino\_wm} 0a9492f; \texttt{miccooper9/egowm} 4a24b6f; \texttt{wzzheng/OccWorld} 1ee7f77\\
\addlinespace
Second (8) & \texttt{autonomousvision/transfuser} 21bdb7e; \texttt{BraveGroup/Drive-WM} eabe3c6; \texttt{hustvl/DiffusionDrive} 9b52ed0; \texttt{jchengai/pluto} b9964b6; \texttt{real-stanford/universal\_manipulation\_interface} d095ba9; \texttt{wayveai/mile} c8b245c; \texttt{shalfun/DriVerse} fc29a19; \texttt{ucd-dare/CarDreamer} 1601324\\
\addlinespace
Third (24 new) & \texttt{H-EmbodVis/HERMESV2} eef5bed; \texttt{autonomousvision/tuplan\_garage} b51d5d0; \texttt{Huster-YZY/GenieDrive} da48a52; \texttt{ucla-mobility/AutoVLA} ba34eed; \texttt{hustvl/Senna} 31a3a23; \texttt{autonomousvision/plant} 7691a93; \texttt{JeffWang987/DriveDreamer} da1ca92; \texttt{vita-epfl/MAD-World-Model-Code} b206702; \texttt{LMD0311/HERMES} 7cb9d79; \texttt{swc-17/SparseDrive} fbadea6; \texttt{dtc111111/GaussianDWM} 50e29cb; \texttt{adept-thu/MomAD} 7d22476; \texttt{SenseTime-FVG/OpenDWM} b0ecc3d; \texttt{xiaomi-research/recogdrive} 5831ee8; \texttt{UWMILab/LCVN} 7318c32; \texttt{leggedrobotics/viplanner} 6fcf3c6; \texttt{dk-liang/UniFuture} 6756cb2; \texttt{wzzheng/GenAD} b16667f; \texttt{Ma-Zhuang/OmniNWM} 8e2333e; \texttt{opendilab/InterFuser} f0be8ea; \texttt{Thinklab-SJTU/Bench2DriveZoo} 498c1f7; \texttt{real-stanford/diffusion\_policy} 5ba07ac; \texttt{vladisai/PLDM} 1bd7e56; \texttt{facebookresearch/jepa-wms} 13cf1d9\\
\bottomrule
\end{tabular}
\end{table*}

\topic{Why most verdicts are weak tests.}
Of the third audit's 11 verdicts (counting those reached after dropping columns), 10 are ``no difference'' results that compare an encoder with our own inverse of the repository's decoder or with one formula shared by both sides; LCVN is the exception. They therefore check little more than whether decoding inverts encoding.

\topic{LCVN at the input level (post hoc).}
LCVN's actor--critic policy learns in rollouts imagined by its world model, and by our reading these rollouts receive actions one row ahead of the world model's training rows. No LCVN world-model or actor--critic checkpoint is public at the audited commit, so we could not measure an effect on predictions or on the policy. We instead read the released validation split (3,602 trajectories, each stored three times with identical data, so we cluster by trajectory) with a restricted unpickler and without network access. Its action field is the world-frame step $p_{t+1}-p_t$ with wrapped yaw change on all 3,602 trajectories, which confirms the convention our audit assumed. With true actions, the translation input the world model receives in actor--critic imagination differs from its training rows for the same frames by a median relative difference of 0.504 [0.501, 0.507] (mean absolute 0.075\,m and 0.079\,rad per row; 95\% bootstrap over trajectories). About a third of this is structural, coming from one extra nonzero row; on the other rows, the median is 0.173 [0.170, 0.176] (exploratory).

\topic{A column rule that failed its check.}
ActRosetta refused LCVN's real columns because one column, the time step, is constant. After seeing this, we wrote a generic rule that drops columns that are constant over the probe paths, locked it with a stop rule, and reran. On LCVN, it named the lag (training lag 2, rollout lag 1) and found no difference in the control pair. On the opened sealed sets, however, it added no false alarm but one wrong name and 49 crashes: on restricted paths, columns can look constant because of the paths rather than the encoder. By our locked stop rule, we do not claim the LCVN naming result.

\section{Command following in a simulator}
\label{app:ctl}
Better images need not mean better decisions. This section tests, outside NWM's training domain, whether repaired actions make predictions follow commands more closely (Sec.~3, ``Where repair matters''); Sec.~\ref{app:loop} reports a decision test and closed-loop tests. We used Habitat~\cite{savva2019habitat} with Matterport3D~\cite{chang2017matterport3d} scenes: 40 cases over 20 scenes, each with 8 candidate action sequences (8 frames at 4\,fps), 50 diffusion steps and one seed. For each candidate, we asked whether NWM's prediction is closest, by LPIPS, to the simulator's real future for that candidate (chance 0.125). The repaired actions here were produced by ActRosetta's adapter from the released outputs. Retrieval was only slightly above chance in both arms (0.172 released, 0.166 repaired), and repair showed no evidence of a change: $-$0.006 [$-$0.066, 0.053] (95\% scene-cluster bootstrap). A matching decision test on the same new scenes first checks that the task can show a benefit at all: with the simulator's real futures in place of predictions (the oracle), choices must clearly beat a fixed always-forward baseline and shuffled futures. This oracle gate failed on these scenes, so by our locked rule we do not interpret the decision result. The earlier five-scene diagnostic of Sec.~\ref{app:loop} passed the same gate. We make no claim about control performance.

\section{Example clips across the range of repair outcomes}
\label{app:grid}
\begin{figure*}[t]
\centering\includegraphics[width=\textwidth]{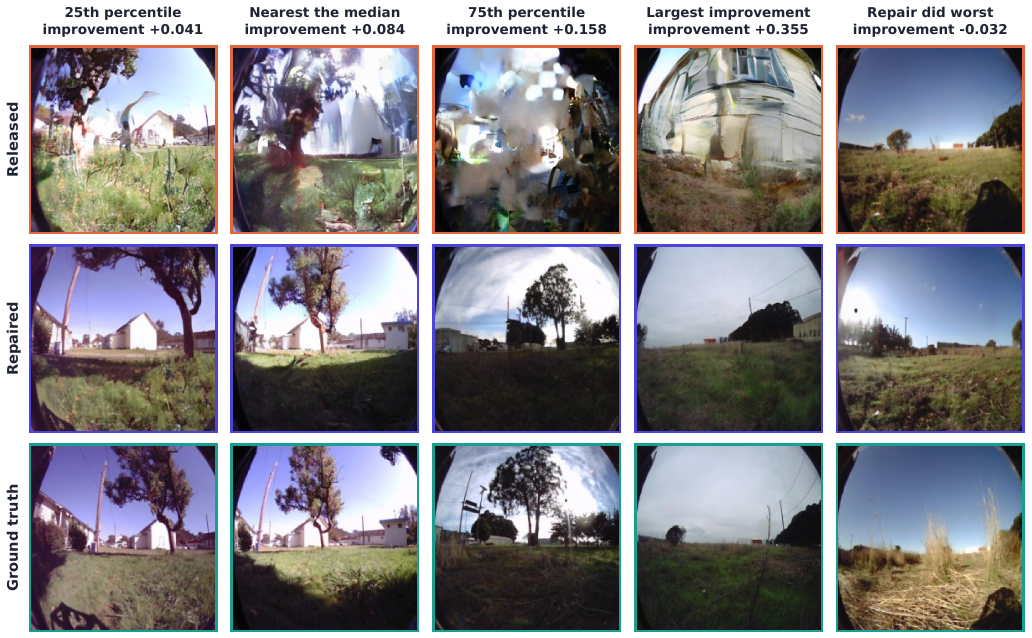}
\caption{\textbf{Examples spanning the range of repair outcomes} (main test: 150 RECON clips, 250 diffusion steps, seed~0). Rows show released predictions, repaired predictions and ground truth at 16\,s. Columns show the clips nearest the 25th, 50th and 75th percentiles of improvement, then the clips with the largest and the smallest (negative) improvement. Improvement is the drop in mean DreamSim error over 1--16\,s from released to repaired actions, not the error of the displayed frame alone; it is used both to select the clips and in the column titles.}
\label{fig:grid}
\end{figure*}

Figure~1 of the main paper shows one RECON clip. Its mean DreamSim improvement over 1--16\,s, 0.083, is close to the median over the 150 main-test clips, 0.085 (49th percentile). Its repaired 16\,s frame, however, matches the ground truth more closely than is typical, so that frame alone overstates the average improvement. Figure~\ref{fig:grid} shows clips across the whole range of outcomes, including one where repair made the prediction worse.

\section{The contract family}
\label{app:family}
Section~4 states Proposition~1 for the core family and points here for its proof sketch and for the weaker evidence for the full family, where lag, stride and aggregation interact. This section gives both, lists the hypotheses the full family contains, and states the recovery step as Algorithm~\ref{alg:recover-supp}.

\begin{proof}[Proof sketch of Proposition~1]
Zero motion gives $b$, and later rows show whether it persists. Two non-collinear translations at heading zero give $M$. A translation at a nonzero start heading separates world coordinates from the other frames. After a turn $\psi\neq0$, first-frame and current-frame displacements differ by $R(\psi)$, and these differ in output because $M$ is invertible. Rotation in place gives the yaw scale and reference. A chunk rotating by more than $\pi$ separates wrapped from unwrapped yaw, and a difference of wrapped headings departs from a wrapped change once the heading crosses the wrap inside a chunk.
\end{proof}

\topic{Hypotheses in the full family.}
Chunk $k$ reads indices $i_k=kms-\ell$ to $j_k=i_k+ms$ (Eq.~1). Position hypotheses combine the frame angle $\alpha_k\in\{\theta_{i_k},\theta_0,0\}$ (current, first, or zero for world coordinates), the anchor $\kappa_k\in\{i_k,0\}$, the aggregation $\mathcal A\in\{$sum, mean, last step$\}$, the lag $\ell\in\{0,\pm1,\pm2\}$, the stride multiple $m\in\{1,2,3,4\}$ and the offset on all chunks or only the first: $3\cdot2\cdot3\cdot5\cdot4\cdot2=720$ hypotheses. Yaw hypotheses combine 7 choices of rule (a wrapped or unwrapped change measured from the chunk start, the sequence start or zero heading, or a difference of wrapped headings), 3 aggregations, 5 lags and 4 stride multiples: $7\cdot3\cdot5\cdot4=420$ hypotheses. The continuous parts are the $2\times2$ map $M$, the offset $b$, a symmetric per-axis bound $C$, the yaw scale and the yaw offset, and $Q$ outputs a scalar, a sine--cosine pair or a cosine--sine pair.

\topic{Evidence for the full family.}
For the full family, we have no proof, only an exhaustive finite check at one setting of the continuous parameters. In it, all 720 position hypotheses and 397 of the 420 yaw hypotheses were separated from every behaviourally different alternative. The other 23 are wrapped-versus-difference pairs at unusual stride and lag whose chunks never cross the wrap on the short probe set; ActRosetta reports them as ambiguous. Other parameter values have only empirical support.

\topic{What naming leaves out, and what the family covers.}
Witness-backed naming never names three kinds of difference: a differing saturation bound, a reference that last-step aggregation makes irrelevant, and a bound that only follows a change of scale. The family is a design choice, not the result of a survey. It represents the encoders of NWM, MWM, RAE-NWM, Vid2World, ViNT, GNM, NoMaD and World-in-World. On the eight new repositories of the first audit, it gave a verdict on three (Sec.~\ref{app:cov}).

\begin{algorithm}[t]
\caption{Recover one encoder's contract (the recovery step of Sec.~5).}
\label{alg:recover-supp}
\footnotesize
\begin{algorithmic}[1]
\Require encoder $E$, probes $B$, family $\Phi$, range paths $L$, hold-out $V$, tolerance $\epsilon$
\If{two identical calls differ} \Return \textsc{unsupported} \EndIf
\State $Y\gets E(B)$
\For{each discrete $h\in\Phi$, position and yaw separately}
  \State fit the continuous part of $h$ to $Y$ by robust least squares
  \State keep $h$ if its residual is at most $\epsilon$
\EndFor
\If{nothing is kept} \Return \textsc{outside family} \EndIf
\State drop $h$ if it misses $E(L\cup V)$ by over $10\epsilon$, unless the misses are a clean saturation at one bound per axis, which sets $C$
\If{nothing is left} \Return \textsc{failed verification} \EndIf
\State refuse if the contract misses fresh paths at higher speed
\State group what is left into behavioural classes on $\mathcal D$
\State find unexcited fields from the rank of the design matrix
\If{one class and no unexcited field} \Return \textsc{definite}, canonical contract
\Else \ \Return \textsc{ambiguous}, the classes, unexcited fields, next probe \EndIf
\end{algorithmic}
\end{algorithm}

\section{Sensitivity of witness-backed naming}
\label{app:sens}
\begin{table}[t]
\centering\footnotesize
\setlength{\tabcolsep}{2.5pt}
\caption{\textbf{Sensitivity of witness-backed naming on the final sealed set} (post hoc, opened set; 226 deterministic pairs, the 14 noisy pairs excluded). One setting is changed at a time from the frozen code. All three tolerances are relative to the output scale. The \emph{witness margin} is the output gap a witness path must show (frozen $10^{-3}$); the \emph{closure tolerance} decides whether a set of field changes makes the evaluation contract behave as the training contract (frozen $10^{-4}$); and the \emph{recovery tolerance} is the residual tolerance $\epsilon$ of Algorithm~\ref{alg:recover-supp} (frozen $10^{-5}$). With the whole naming rule removed, ActRosetta names every field that differs for every pair of accepted contracts. Exact recovery is out of the 91 differing pairs; wrong names are fields named outside a pair's true differences. In every setting, refusal stayed at 48 of 48 and no false alarm occurred. Encoder calls are deterministic, so no seeds apply.}
\label{tab:sens}
\begin{tabular}{@{}p{.47\linewidth}cc@{}}
\toprule
Setting & Exact recovery & Wrong names\\
\midrule
Frozen code (witness margin $10^{-3}$) & 73/91 & 1\\
Whole naming rule removed & 68/91 & 4\\
Witness requirement removed & 73/91 & 2\\
Witness margin $10^{-4}$ or $3\times10^{-4}$ & 73/91 & 2\\
Witness margin $3\times10^{-3}$ & 72/91 & 1\\
Witness margin $10^{-2}$ & 68/91 & 1\\
Closure tolerance $3\times10^{-5}$ or $3\times10^{-4}$ & 73/91 & 1\\
Recovery tolerance $3\times10^{-6}$ or $3\times10^{-5}$ & 73/91 & 1\\
Recovery tolerance $10^{-4}$ & 73/91 & 2\\
Five rigid transformations of the probes & 70, 73, 73, 72, 70 & 1 each\\
\bottomrule
\end{tabular}
\end{table}
Witness-backed naming has a few settings that we fixed before the final run. To show how much the final result depends on them, we varied one setting at a time on the opened final set (Table~\ref{tab:sens}); these numbers show fragility and are not held-out evidence. Removing the whole naming rule lowers exact recovery from 73 to 68 of 91 and raises wrong names from 1 to 4. Removing only the witness requirement keeps recovery at 73 but doubles the wrong names to 2. A smaller witness margin also admits a second wrong name, and a larger one costs up to five exact recoveries. The two tolerances leave recovery at 73 and add one wrong name only at the loosest recovery tolerance, and rigid transformations of the probe paths move recovery between 70 and 73. Across all settings, exact recovery stays between 68 and 73 of 91 and wrong names between 1 and 4; the witness requirement is what removes the second wrong name.

\section{Sealed runs: scoring rules and final-run details}
\label{app:history}
\begin{figure*}[t]
\centering\includegraphics[width=\textwidth]{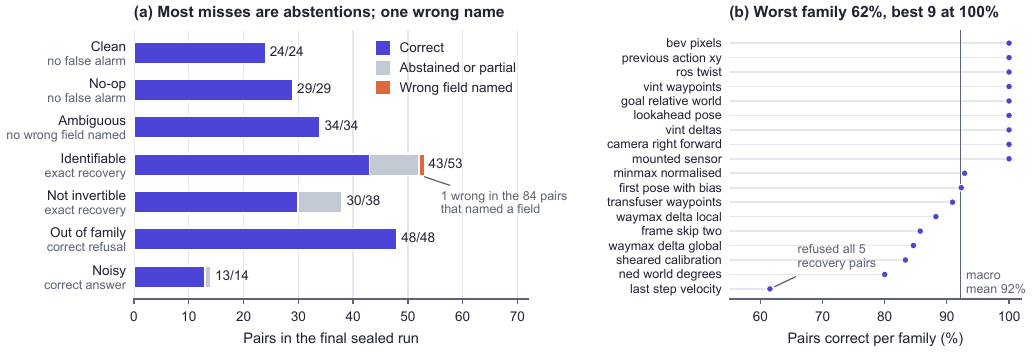}
\caption{\textbf{ActRosetta on the final sealed set: 240 pairs from 18 interface families.} (a) Pairs per category, each scored by the criterion under its name. Gray marks pairs that were refused, abstained on or named in part; orange marks a wrong field. Identifiable and non-invertible pairs are the 91 differing pairs. (b) Share of pairs correct in each interface family; the vertical line is the unweighted mean. ``Recovery pairs'' in the panel means differing pairs.}
\label{fig:sealed-supp}
\end{figure*}

Section~6 describes the sealed runs. This section records how the protocol and its scoring rules changed between runs, and gives details of the final run (Figure~\ref{fig:sealed-supp}).

\topic{Order of the runs.}
Runs 1--3 were development challenges: after each, we traced the misses, fixed the code and froze it by hash again before the next set was written. We originally declared the third run the last. We then added witness-backed naming, which we had developed on the three opened sets and which therefore needed a held-out test, so we ran a fourth, final run. All four results stand as scored (Table~2). Exact recovery did not rise steadily (92\%, 81\%, 88\% and 80\%), so the runs should not be read as steady progress.

\topic{Changes to the scoring rules.}
Two scoring rules changed, each before the next set existed. From run~2 on, a pair counts as refused if either of its encoders is refused. From run~3 on, a definite ``no difference'' on an ambiguous pair's restricted paths is not an error (Table~2, footnote).

\topic{Final-run details.}
In the final run, the one wrong name concerns a general $2\times2$ map that ActRosetta calls ``axis'' and the label calls ``scale''; it remains counted as wrong. Results on noisy pairs are in Secs.~\ref{app:base} and~\ref{app:probe}.

\section{Settings and intervals for Section 3}
\label{app:moved}
This section gives the settings behind Figures~2--4 and the intervals that Section~3 omits.

\topic{Settings.}
Figure~2b reports 95\% trajectory-level confidence intervals for the main test and 99\% day-level intervals, corrected for five tests, for the new cohort. Figure~2a and the main-test row of Figure~2b use 150 RECON clips from 20 trajectories, 16\,s at 1\,fps, 250 diffusion steps and seed~0. The new cohort in Figures~2b and~3 uses 120 clips from 60 trajectories recorded on 20 days, three seeds, 16\,s at 1\,fps and 250 diffusion steps. Figure~2c uses 50 clips from 15 trajectories, 50 diffusion steps and seed~0, over 16\,s: 16 predicted frames at 1\,fps and 64 at 4\,fps.

\topic{Rollout error.}
Across three noise seeds at 50 diffusion steps (150 main-test clips), repair reduces mean DreamSim error by 0.098 [0.073, 0.119]; at 4\,fps, on 50 clips at 50 diffusion steps, the reduction is 0.075 [0.060, 0.099]. The released script's \texttt{torch.compile}, which our generation code omits, lowers released 16\,s DreamSim error by 0.014 [0.004, 0.026] (unrounded in Sec.~\ref{app:anchor}) and leaves the 16\,s reduction almost unchanged: 0.198 [0.147, 0.231] with compile on, against 0.203 with compile off, on the 74 clips run both ways. The S checkpoint has an action-responsiveness ratio similar to the Ego4D checkpoint's (1.37 against 1.40) and gains ten times as much from repair (Sec.~\ref{app:faults}).

\topic{Checkpoint ranking.}
On the new cohort (paired day-level bootstrap), the DreamSim gap between L and the Ego4D checkpoint, L minus Ego4D, goes from +0.044 [0.025, 0.065] with released actions to $-$0.048 [$-$0.055, $-$0.042] with repaired ones. Kendall's $\tau$ between the released and repaired orders is 0.2 under both DreamSim and LPIPS. The four pairwise orderings that reverse, with both intervals excluding zero under both metrics, are XL--L, XL--Ego4D, B--Ego4D and L--Ego4D; after Holm correction~\cite{holm1979simple} over the ten pairs, the change between encodings has adjusted $p=0.001$ for each, the smallest value the 10,000-resample bootstrap can give. When any one of the 20 recording days is left out, the Ego4D checkpoint stays first with released actions and L first with repaired actions in all 20 cases.

\begin{table*}[t]
\centering\footnotesize
\setlength{\tabcolsep}{3pt}
\caption{\textbf{Alternative models' lead over NWM on 90 RECON clips, with NWM's released and repaired actions.} Lead is NWM's mean error minus the alternative model's, as in Figure~4: positive means the alternative model has lower error. 90 clips from 19 trajectories, 16 frames at 4\,fps (4\,s), seeds 0--4 averaged within each clip, all models on RTX 3090 Ti GPUs (one GPU type), each at its own settings (NWM: XL, 250 diffusion steps). Brackets: 95\% trajectory-cluster bootstrap intervals (10,000 resamples); parentheses: Holm-corrected $p$ over two models and two metrics within each action condition. Last column: DreamSim lead with repaired actions at each seed.}
\label{tab:cross}
\begin{tabular}{@{}lllll@{}}
\toprule
Metric & Model & Released actions & Repaired actions & Repaired, by seed (0--4)\\
\midrule
DreamSim & EgoWM & +0.0101 [+0.0016, +0.0213] (0.014) & $-$0.0108 [$-$0.0148, $-$0.0048] (0.007) & $-$0.0076, $-$0.0127, $-$0.0128, $-$0.0117, $-$0.0092\\
DreamSim & RAE-NWM & +0.0197 [+0.0134, +0.0271] (0.001) & $-$0.0012 [$-$0.0074, +0.0048] (1.0) & +0.0012, $-$0.0017, $-$0.0028, $-$0.0017, $-$0.0012\\
LPIPS & EgoWM & +0.0412 [+0.0286, +0.0570] (0.001) & +0.0016 [$-$0.0070, +0.0111] (1.0) & \\
LPIPS & RAE-NWM & +0.0349 [+0.0260, +0.0462] (0.001) & $-$0.0047 [$-$0.0153, +0.0046] (0.99) & \\
\bottomrule
\end{tabular}
\end{table*}

\topic{Comparison with alternative models.}
Table~\ref{tab:cross} gives the comparison behind Figure~4. Its 90 clips are the 30 clips of an earlier comparison (below) and 60 drawn from a seeded permutation of the remaining 120 main-test clips; the clip count was fixed by a rule locked before any new score. The horizon, 4\,s, is the longest that EgoWM's own script supports. All models are scored on a shared image view with the same scorer, and each model's per-clip error is averaged over the five seeds before bootstrapping. By the locked rule, repair removes EgoWM's lead, and RAE-NWM's lead shrinks from 0.0197 to $-$0.0012, which is no detectable difference, not equivalence. Every seed alone points the same way for EgoWM, and RAE-NWM's per-seed leads stay within 0.003 of zero. On the 60 new clips alone (12 trajectories), EgoWM's lead with released actions was 0.0039 [$-$0.0028, +0.0141], so there was no clear EgoWM lead to begin with; with repaired actions it was $-$0.0115 [$-$0.0163, $-$0.0037]. Mean DreamSim errors are 0.1362 for NWM with released actions, 0.1153 for NWM with repaired actions, 0.1261 for EgoWM and 0.1165 for RAE-NWM. Between the RTX 3090 Ti and the T4 and L4 GPUs of the earlier comparison (same clip, seed and inputs), RAE-NWM's per-clip DreamSim error moved by a mean absolute 0.006 to 0.008 (maximum 0.035), close to the size of the repaired differences, while EgoWM's moved by $10^{-4}$ or less; this is why all models ran on one GPU type. The comparison is limited to RECON, 19 trajectories and 4\,s; the 30 earlier clips had been seen before at other seeds and on other machines; and EgoWM uses 1 context frame, the others 4.

\topic{Earlier 30-clip comparison.}
Before the comparison above, we compared the models on 30 RECON clips (19 trajectories, 4\,s at 4\,fps), each model at its own settings, with the models run on different GPU types. Because RAE-NWM's outputs change across GPU types, this comparison was superseded for EgoWM and RAE-NWM; it remains the only source of the Vid2World result. Over three seeds, EgoWM and RAE-NWM led NWM with released actions by DreamSim +0.021 [+0.007, +0.040] and +0.024 [+0.012, +0.040]. With repaired actions, their leads became $-$0.010 [$-$0.016, $-$0.004] and $-$0.008 [$-$0.014, $-$0.001], with the same sign at every seed. By LPIPS, RAE-NWM's lead also reversed ($-$0.014 [$-$0.025, $-$0.004]), while EgoWM's interval included zero ($-$0.009 [$-$0.019, +0.002]). On the first seed alone, both DreamSim intervals included zero. Vid2World, run with one seed, kept a clear but smaller lead: +0.022 [+0.017, +0.027] with repaired actions, from +0.054 [+0.038, +0.073] with released actions.

\section{Noisy logs, active probing and cost}
\label{app:probe}
This section describes two optional modes of ActRosetta, for noisy logs and for active probing, and its computational cost (Sec.~5). On the development interfaces, with declared noisy logs, ActRosetta answered 60--77\% of 120 cases with no wrong answer (Sec.~\ref{app:base}), and across the four sealed runs it was correct on 52 of 54 noisy pairs, with one wrong answer. On the development cases, the answers were also all correct without the stability gate defined below, so the benefit of the gate is not shown.

\topic{Noisy logs.}
Exact tolerances are the wrong tool when logged poses differ from those the encoder saw. In noisy mode, which the user declares, ActRosetta refits on five perturbed probe sets. It answers only if the same hypothesis wins every time and the next behaviourally different hypothesis has at least 5 times the residual; we call these two conditions the \emph{stability gate}, and when either fails it reports \textsc{unsupported}. If the best residual exceeds 2\% of the output scale, it reports the encoder as outside the family. These values were fixed before the sealed tests.

\topic{Counterexample-guided probing.}
The active mode starts from one path. At each step, it queries the path from a pool that best separates the classes still accepted, or the probe kind linked to an unexcited field, and refits. Without noise, each query with a separating path removes at least one class, so for a finite family whose classes are pairwise separable on the pool, it ends with the right class after at most one query fewer than the number of classes. ActRosetta also returns each recovered contract as a YAML file, and witnesses can serve as unit tests.

\topic{Cost.}
Position and yaw are fitted separately, so one recovery solves $720+420$ small least-squares problems, with at most 3 unknowns per output, instead of a joint search over $720\times420=302{,}400$ hypotheses. Cost grows linearly with the number of hypotheses per part, and recovery needs no model weights or GPU. On the final sealed set, ActRosetta took a median of 0.58\,s and 400 transitions per pair.

\section{Baseline setup and real-code protocol}
\label{app:proto}
Sections~6 and~7 compare ActRosetta with baselines on the final sealed set (Table~3) and test it on real training--evaluation pairs. This section gives the baselines' settings and the protocol for the real-code pairs (Table~\ref{tab:real-supp}).

\topic{Baseline setup.}
Random fuzzing calls both encoders on random paths (uniformly drawn per-step motions from a random start pose) at chunk sizes 1, 2 and 4, and reports a difference when an output differs by more than $10^{-6}$, changes shape or raises an exception. The affine adapter (least squares) and an MLP adapter (two hidden layers of 64 ReLU units, trained with Adam) are fitted to map evaluation outputs to training outputs on the same budget of 280 transitions per pair. Each adapter sees the current evaluation row, the three previous rows and the chunk index, and is trained on paired evaluation and training outputs; ActRosetta needs no paired data. The baselines ran untuned, after the final set was opened. We did not measure the time a manual audit would take.

\begin{table}[t]
\centering\footnotesize
\setlength{\tabcolsep}{2.5pt}
\caption{\textbf{Real training--evaluation pairs under witness-backed naming} (Sec.~7), with expected results written from code reading before running. F: reference frame; O: normalization offset; W: yaw wrapping; unres.: unresolved. Executed: the repository's own function ran; Static: compared by reading. Same: one behavioural class; No diff.: no observed difference. NWM informed development. World-in-World (WiW) is tested only on paths built from its discrete commands (forward 0.2\,m, turn 22.5$^\circ$, stop); its corrected encoder is our patch. \icok\ matches the expected result, \icpart\ matches it in part.}
\label{tab:real-supp}
\begin{tabular}{p{.42\linewidth}p{.14\linewidth}cc}
\toprule
Training vs.\ evaluation & Status & Expected & Named\\
\midrule
NWM vs.\ NWM rollout & Executed & F, O, W & F, O, W \icok\\
NWM vs.\ MWM rollout & Executed & F, O, W & F, O, W \icok\\
NWM vs.\ CompACT rollout & Static & F, O, W & F, O, W \icok\\
NWM vs.\ WiW released & Executed & F, O, W & F, O; W unres. \icpart\\
RAE-NWM vs.\ rollout & Executed$^\ddagger$ & None & Same \icok\\
NWM vs.\ Vid2World rollout & Executed & None & Same \icok\\
NWM vs.\ WiW corrected & Our patch & None & No diff. \icok\\
\bottomrule
\end{tabular}

\smallskip{\footnotesize $^\ddagger$The rollout composition was transcribed by hand.}
\end{table}

\topic{Real-code protocol.}
For each real pair, we wrote the expected diagnosis from code reading before running ActRosetta. The encoders of NWM, MWM, RAE-NWM, Vid2World, ViNT and World-in-World ran as the repositories' own functions at pinned commits, on synthetic paths (ViNT is a development interface, Sec.~\ref{app:base}, not a row of Table~\ref{tab:real-supp}). The last three were first transcribed by hand and then matched against the executed functions: exactly for Vid2World and World-in-World, and to $3.05\times10^{-6}$ for ViNT, whose code casts to float32. CompACT~\cite{kim2026planning} was compared with NWM by reading only. The audited commits are \texttt{3f6cd8e} for NWM, \texttt{e5a5429} for MWM, \texttt{71b3029} for CompACT (its only public commit), \texttt{1464a67} for RAE-NWM, \texttt{ae24077} for Vid2World, \texttt{6ac81ef} for World-in-World and \texttt{dca7981} for ViNT (the visualnav-transformer repository).

\topic{Repair from NWM's real outputs.}
On 20 real RECON clips, the path that the unrepaired NWM rollout's actions command ends a median 7.57\,m from the recorded endpoint, while the adapter, applied to those rollout outputs alone, reproduces the training tensors to $5.7\times10^{-15}$ (Sec.~7).

\section{NWM's planning code}
\label{app:plan}
Section~3 states that NWM's planning code builds actions by a separate route that matches training at default settings. This section gives that check, a separate difference in how plans are scored, and an earlier planning test. All checks are on CPU, without the model, unless stated.

\topic{What the model receives.}
At the audited commit, planning does not use the mismatched rollout action encoder: \texttt{planning\_eval.py} reads \texttt{TrajectoryEvalDataset}, and a plan is one normalized per-step displacement repeated 8 times. We ran its action builder on 500 plans drawn from the initial distribution of the cross-entropy method (CEM)~\cite{rubinstein1999cross} planner on RECON (seed 20261105) and compared the result with the training encoder. At stride~1, the setting in the repository's README, the actions the model receives match training to floating-point precision, and ActRosetta reports no observed difference. Strides 2 and 4 sum normalized raw-step displacements, so the offset accumulates (largest row difference in $x$: 0.333 and 1.000). The README does not use these strides, but a user who sets them gets a real mismatch.

\topic{How plans are scored.}
Planning then scores each plan's absolute trajectory error (ATE) and relative pose error (RPE) along a straight line (\texttt{get\_action\_torch}), not along the path the model imagines. At stride~1, the largest row differences between the two paths are 0.205 ($x$) and 0.353 ($y$), and ActRosetta names \emph{frame} on the scored path, with a witness of 8 transitions. This concerns how plans are scored, not what the model receives. A geometric analysis with no planner run (100 RECON start states from 36 trajectories, goal 2\,s ahead, 95\% trajectory bootstrap) shows its size. For the plan whose imagined endpoint reaches the goal, straight-line scoring gives an ATE of 0.79 [0.70, 0.87], against 0.50 [0.42, 0.57] along the imagined path, a gap of +0.29 [0.24, 0.35] waypoint units (0.07\,m), and RPE rises by +0.10 [0.09, 0.12]. Over 120 random candidates per state, ATE rankings under the two scorings agree with Kendall's $\tau$ of 0.76 [0.75, 0.78]. CEM ranks candidates by LPIPS, not ATE, so these ranking changes do not act on the planner directly. For near-ideal plans this scoring would penalize NWM's ATE, but real CEM plans are not ideal, so its effect on NWM's planning table is unmeasured. The NWM + NoMaD rows of that table cannot be audited, as their code is not in the repository.

\topic{An earlier planning test.}
An earlier offline planning test with the model (20 cases from 18 trajectory clusters, 20 candidates, three seeds, 250 diffusion steps, 2,400 rollouts) found no planning effect of repaired actions: released minus repaired endpoint error was $-$0.0077\,m (95\% trajectory $t$ interval [$-$0.0748, +0.0594]). It is a reduced-candidate offline diagnostic, not closed-loop navigation or a reproduction of the published table.

\section{Closed-loop, decision and history tests}
\label{app:loop}
Section~3 reports that repair does not measurably improve command following (40 cases; Sec.~\ref{app:ctl}) or image-goal decisions (20 cases, two runs). This section gives the decision test, a closed-loop test in World-in-World~\cite{zhang2026world}, which places NWM inside tasks where its predictions guide an agent, and two tests of the context frames. Together they show no measurable benefit of repair for decisions, at sample sizes that cannot exclude small effects.

\topic{Image-goal decisions.}
This diagnostic in Habitat with Matterport3D scenes (20 cases in five scenes, eight fixed candidates of 8 transitions each, an LPIPS score on the final image, and \emph{regret}: the geodesic distance from the chosen candidate's endpoint to the goal minus the smallest such distance among the eight candidates, in metres) first had to pass an oracle gate: with the simulator's real futures in place of predictions, choices must beat simple baselines. On these five scenes it passed: regret fell by 0.4962\,m [0.4271, 0.6107] against a fixed always-forward baseline and by 0.9300\,m [0.6765, 1.1834] against shuffled futures. (On the 20 new scenes of Sec.~\ref{app:ctl}, the same gate failed.) With NWM's predictions (960 rollouts per run, three paired seeds, 50 diffusion steps), repair did not reduce regret: released minus repaired was $-$0.1197\,m [$-$0.2352, $-$0.0006] in the first run and $-$0.1487\,m [$-$0.3035, 0.0213] in a replication with new cases on the same scenes (95\% scene-cluster bootstrap). Both point estimates favour released actions, and the first run's bootstrap interval lies just below zero. With only five scene clusters, however, the scene-level $t$ intervals ([$-$0.3086, 0.0691] and [$-$0.4123, 0.1149]) include zero, and the replication's includes the +0.10\,m effect we had set in advance as useful. We therefore claim neither a benefit nor a general harm of repair in this diagnostic. Collisions, repeated stationary context frames and the simulator's domain shift limit what this says about the action interface. A pipeline check found that the decision loop's rollouts equal those of the released generator exactly (18 of 18 tensors).

\topic{World-in-World, Active Recognition.}
World-in-World's conversion of NWM actions drops the normalization offset (Table~\ref{tab:real-supp}). On the first 100 episodes of a seeded permutation of its 551 Active Recognition episodes (26 scenes), with NWM at 50 diffusion steps, seed~0, and a Qwen2.5-VL-7B~\cite{bai2025qwen25vl} planner at temperature~0, success was 28 of 100 with World-in-World's actions and 28 of 100 with repaired ones; the planner alone, without NWM, scored 31 of 100. No episode differed in outcome (exact McNemar $p=1.0$). The scene-cluster bootstrap interval of the difference is degenerate ([0.000, 0.000]); if episodes are treated as independent, 0 discordant episodes of 100 exclude a change above about 3.6 percentage points (95\% Clopper--Pearson). This is no evidence of a change at this sample size, which is not evidence of no effect. World-in-World runs NWM at 50 rather than 250 diffusion steps; on one episode with ten calls, the DreamSim difference between 50 and 250 diffusion steps was no larger than that between two 250-step samples with different noise ($-$0.028 [$-$0.068, +0.006]).

\topic{History of the context frames.}
NWM predicts from four context frames, and in the decision loop these were four copies of the latest frame. On the same 20 image-goal cases with repaired actions (960 rollouts, three seeds, 50 diffusion steps), a genuine history of three 30$^\circ$ turns in place of the copies did not improve the final prediction: the LPIPS reduction was $-$0.0008 [$-$0.0044, 0.0028], and the regret reduction $-$0.0080\,m [$-$0.1305, 0.1133], an interval that includes useful gains (95\% scene-cluster bootstrap). On 50 RECON clips (20 trajectories, three seeds, 50 diffusion steps), a moving recorded history in place of a repeated frame did not change DreamSim error: the reduction was $-$0.0005 [$-$0.0044, 0.0034]. This interval is a 98.3\% $t$ interval over trajectory means, Bonferroni-adjusted for the three tests run as one locked suite: this one, the robustness check of Sec.~\ref{app:more} and the planning test of Sec.~\ref{app:plan}. Motion and visual coverage changed together, so these tests do not separate them.

\section{Additional robustness checks and negative results}
\label{app:more}
This section reports checks that support Sec.~3 but that its claims do not rely on, and two directions that did not work.

\topic{Repair on previously examined trajectories.}
On 50 RECON clips from 20 previously examined trajectories (three seeds, 250 diffusion steps), repair lowered mean DreamSim error from 0.2324 to 0.1375, a reduction of 0.0949 [0.0554, 0.1343] (98.3\% $t$ interval over trajectory means, Bonferroni-adjusted for the three tests of the same locked suite; Sec.~\ref{app:loop}). These trajectories had been examined before, so this is a robustness check, not an independent replication.

\topic{MWM on SCAND.}
MWM~\cite{yan2026mwm} builds on NWM and shares its rollout mismatches (Sec.~7). On NWM's SCAND split (150 clips, 13 trajectories, 1\,fps, 16\,s, seed~0), MWM, run from its released LoRA weights~\cite{hu2021lora} at 5 diffusion steps, leads NWM at 5 diffusion steps by DreamSim +0.179 [+0.139, +0.207] with released actions and +0.166 [+0.125, +0.199] with repaired ones. The mismatch thus inflated MWM's advantage by +0.012 [+0.002, +0.024], about 7\%, and the advantage survives. MWM's paper also reports that MWM at 5 diffusion steps beats NWM at 25; with the released weights, they tie on DreamSim ($-$0.003 [$-$0.024, +0.013] with released actions), and NWM at 25 diffusion steps is better on LPIPS by 0.063 [0.053, 0.073]. Possible causes are untested: the released LoRA weights are rank 16 on all linear layers, unlike the paper's description; our SCAND copy and resolution may differ; and MWM's evaluation samples without a fixed seed. This is therefore a reproducibility question, not a claim of error.

\topic{Vid2World pilot.}
On 10 RECON clips (9 trajectories, 50 diffusion steps, 16 frames at 4\,fps; seeds 0 and 1 for the clip's own actions, seed 0 for the substituted ones), substituting another clip's actions changed Vid2World's frames 2.308 [1.933, 2.752] times as much as a new seed did (DreamSim), and more on all 10 clips. This was a descriptive pilot; the locked 12-clip version is in Sec.~\ref{app:v2w}.

\topic{A learned pose estimator.}
Before the inverse dynamics model and feature matching of Sec.~\ref{app:fid}, we tried to measure motion in predictions with VGGT-1B~\cite{wang2025vggt}, testing it first on ground-truth frames (50 clips per dataset, 800 frame transitions, 95\% trajectory bootstrap). It failed its locked gate on both datasets: the median per-step yaw error was 13.2$^\circ$ [10.6, 15.8] on RECON and 14.7$^\circ$ [12.5, 17.8] on Go Stanford, against a limit of 5$^\circ$, and the yaw slope was 0.378 [0.232, 0.592] and 0.284 [0.154, 0.416], against a required interval inside [0.8, 1.2]. Six variants with denser frames or joint multi-frame windows (groups of frames processed together) also failed, so we did not use this route.

\topic{Uncertainty from sample spread.}
We also tested whether the spread of sampled futures (8 per clip, 50 clips per dataset) indicates prediction error. Spread predicted error at the same prediction time across clips (AUROC 0.71--0.91 on RECON and 0.53--0.77 on Go Stanford), but not later error (AUROC 0.50--0.57). In a ranking test with a shared start (30 RECON start states, 16 candidates each), adding spread to the cost never helped: cross-validation chose zero weight in every fold. We computed no intervals for these AUROCs. This direction was negative.

\section{Why a wrong encoding hurts}
\label{app:mech}
Section~3 reports that the path-preserving $2\pi$ fault raises NWM's error as much as the released actions do. This section asks why a wrong encoding hurts: through the wrong path it commands, through inputs outside the range seen in training, or both. The study is exploratory: each protocol was locked before its generation but designed after the main results. Unless stated, all runs used one H100, the first 50 main-test clips (15 trajectories), 50 diffusion steps, seed~0, 1\,fps and 16\,s; repeated rollouts were identical, and released minus repaired error was 0.1008 DreamSim, the same as on the RTX 3090 Ti. Intervals are 95\% bootstrap intervals over trajectories (10,000 resamples). Values are DreamSim error of each arm minus that of repaired actions. We call an action row \emph{out of range} when its Mahalanobis distance from the training rows exceeds their 99th percentile.

\begin{table*}[t]
\centering\small
\setlength{\tabcolsep}{5pt}
\caption{\textbf{Injected faults raise EgoWM's and RAE-NWM's error.} Increase in mean DreamSim error over 16 predicted frames (4\,s at 4\,fps) relative to correct actions, on 90 RECON clips from 19 trajectories, each model at its own settings. Brackets: 95\% trajectory-cluster bootstrap intervals. Every entry is positive and passes Holm correction over the eight tests of its column ($p=0.0016$ each). Combined is frame plus offset, because the natural yaw mismatch does not occur on these clips. ``Redefined'': adapted to RAE-NWM's own convention (Sec.~\ref{app:faultx}).}
\label{tab:faultx}
\begin{tabular}{llcc}
\toprule
Model & Injected fault & Seed 0, RTX 3090 Ti & Seeds 1--2 mean, A100\\
\midrule
EgoWM & Frame & 0.0222 [0.0139, 0.0305] & 0.0209 [0.0133, 0.0279]\\
EgoWM & Offset & 0.0182 [0.0121, 0.0255] & 0.0184 [0.0113, 0.0271]\\
EgoWM & Combined & 0.0292 [0.0238, 0.0351] & 0.0266 [0.0218, 0.0311]\\
EgoWM & $2\pi$ yaw & 0.1800 [0.1638, 0.1996] & 0.1804 [0.1640, 0.2002]\\
RAE-NWM & Frame & 0.0307 [0.0241, 0.0376] & 0.0312 [0.0246, 0.0378]\\
RAE-NWM & Offset (redefined) & 0.0193 [0.0134, 0.0261] & 0.0189 [0.0134, 0.0255]\\
RAE-NWM & Combined & 0.0333 [0.0262, 0.0398] & 0.0355 [0.0291, 0.0415]\\
RAE-NWM & $2\pi$ yaw (redefined) & 0.0920 [0.0819, 0.1023] & 0.0926 [0.0802, 0.1058]\\
\bottomrule
\end{tabular}
\end{table*}
\topic{Out-of-range input with the true path.}
Adding $\pm2\pi$ to the yaw of every chunk after the first leaves the commanded path identical to the true one and puts every yaw row far outside the training range. Error rose by 0.128 [0.106, 0.162], and it rose on all 50 clips. This is an extreme synthetic intervention. The released yaw mismatch adds $2\pi$ jumps to only some rows: added to otherwise correct rows, those jumps raise error by 0.019 [0.012, 0.031] on the 25 clips that have them. A few out-of-range yaw rows thus do little, and out-of-range yaw in every row does a lot. This fits the small effect of correcting only yaw (Sec.~\ref{app:faults}), which removes a few jumps from rows whose damage is dominated by the frame and offset mismatches.

\topic{One mismatch at a time.}
For each mismatch, we blended released into repaired rows at doses 0.25, 0.5, 0.75 and 1. At dose~1, the frame mismatch raises error by 0.150 [0.107, 0.176], the offset mismatch by 0.006 [$-$0.002, +0.015], and all three together by 0.101 [0.076, 0.120]; yaw raises it by 0.019 on its 25 clips. Under DreamSim, the frame mismatch's damage grows faster at higher doses; under LPIPS, that curvature is not clear. The mismatches do not add up: the sum of the three single-mismatch damages, with yaw averaged over all 50 clips, exceeds the combined damage by 0.065 [0.041, 0.081].

\topic{Pulling inputs back into range.}
Projecting released rows into the training range, in whitened coordinates, cuts error by 0.052 [0.012, 0.075]. It also moves the commanded path, however: the share of the path error it removes (0.26) lies within the interval of the share of the released-action damage it removes (0.52 [0.15, 0.67]), so the two effects cannot be separated. A simple per-axis clip helps as much while leaving 79\% of rows out of range, so being in range is not the whole story. The projection leaves the error of repaired rows unchanged (+0.0001).

\topic{Regression over all arms.}
Over 15,750 chunk-level observations from all arms of this study, damage rises both with the commanded path's error and with the accumulated out-of-range score, each with the other held fixed (standardized coefficients 0.50 [0.47, 0.54] and 0.36 [0.33, 0.42]; $R^2$ 0.38). This is a regression on our designed arms, not a causal decomposition, but it indicates that both mechanisms act in NWM.

\topic{Predicting damage in advance.}
A simple model fitted on 29 existing NWM intervention arms (on an RTX 3090 Ti) predicted the damage of 5 new combinations of two injected faults, with the predictions hashed before generation. Its mean absolute error was 0.105, against 0.146 for predicting the mean, but it underpredicted one combination badly (0.51 measured, 0.17 predicted), and its main predictor was chosen after seeing the fit data. With five combinations, this is weak evidence.

\section{Fault injection on EgoWM and RAE-NWM}
\label{app:faultx}
Section~3 tests whether sensitivity to action encoding extends beyond NWM by injecting faults into EgoWM~\cite{bagchi2026walk} and RAE-NWM~\cite{zhang2026rae}. This section gives the design, the fault definitions, the full results (Table~\ref{tab:faultx}) and each model's action responsiveness. The injected faults are expressed in each model's own action convention; they are not bugs in either model's released code.

\topic{Design.}
We fixed the protocol before generation, and the addition of seeds 1--2 before their generation and regardless of the seed-0 results (recorded locally, not publicly registered). Automated agents ran the study, and no one outside the team replicated it. We use the 90 RECON clips of Sec.~\ref{app:moved} (19 trajectories) and 16 generated frames at 4\,fps, with each model's settings unchanged. EgoWM uses one context frame and RAE-NWM four. Seed~0 ran on one RTX 3090 Ti; seeds 1--2 ran on A100 GPUs, with correct actions regenerated there. Each fault-minus-correct contrast pairs the same clip, seed and GPU type.

\topic{Fault definitions.}
EgoWM's inference action convention matches NWM's training convention, so the frame and offset faults map directly. RAE-NWM already rotates displacements, normalizes without an offset and wraps yaw. Its offset fault is therefore redefined as the same physical perturbation, 0.3125\,m of extra forward motion per step after the first, and its $2\pi$ fault is applied at the model input after its own wrapping. No clip crosses the yaw-wrap boundary within 4\,s, so the combined fault is frame plus offset, and NWM's natural yaw mismatch is not tested. The $2\pi$ fault changes the representation but preserves the commanded path.

\topic{Analysis and results.}
DreamSim is the primary metric and LPIPS (AlexNet features) the secondary, both computed on the same shared image view and with the same scorer as the comparison in Sec.~\ref{app:moved}. For the A100 column we average seeds within clips, then bootstrap trajectories with 10,000 resamples (seed 20261007), using the ratio of sums. Holm correction covers eight tests (two models, four faults), separately for each metric and GPU type. Every injected fault harms image accuracy on each GPU type (Table~\ref{tab:faultx}), and every seed alone agrees in direction. LPIPS confirms all eight, also with corrected $p=0.0016$; its combined-fault increases are 0.0610 [0.0507, 0.0725] for EgoWM and 0.0792 [0.0689, 0.0897] for RAE-NWM at seed~0, and 0.0588 [0.0502, 0.0689] and 0.0825 [0.0750, 0.0911] on the A100s. The $2\pi$ fault causes the largest damage in both models even though it leaves the commanded path unchanged. This supports sensitivity to how actions are represented, not a claim about control success.

\topic{Responsiveness.}
With $E$, $F$ and $R=\bar E/\bar F$ defined as in Sec.~\ref{app:faults}, at seed~0, the DreamSim ratio is 2.410 [2.180, 2.718] for EgoWM ($\bar E=0.1624$, $\bar F=0.0674$) and 1.844 [1.650, 2.081] for RAE-NWM ($\bar E=0.1489$, $\bar F=0.0808$). Both lower bounds exceed the 1.5 threshold fixed in advance. Unlike the NWM and Vid2World ratios of Secs.~\ref{app:faults} and~\ref{app:v2w}, these use one seed for the substituted actions and one seed pair for noise. Across NWM, Vid2World, EgoWM and RAE-NWM, the protocol's descriptive comparison of responsiveness (2.25, 2.54, 2.410 and 1.844) with harm (0.0209, 0.0586, 0.0292 and 0.0333) gives a Spearman rank correlation of 0.4. Here NWM's harm is its released-minus-repaired difference from Sec.~\ref{app:moved} and Vid2World's is its missing-offset fault from Sec.~\ref{app:v2w}. Different clips, horizons and fault definitions preclude any claim of a general relationship.

\topic{Checks and limitations.}
All required arms have finite 16-frame scores on all 90 clips. Seed-0 predictions with correct actions reuse earlier outputs from the same machine only after exact pixel reproduction, and their recomputed scores match exactly; repeated A100 clips also match exactly. Independent arithmetic and cluster-bootstrap checks reproduce the fault contrasts and responsiveness intervals. With correct actions and the same seed, the mean absolute DreamSim difference between the RTX 3090 Ti and the A100s is about 0.0001 for EgoWM and 0.0068--0.0070 for RAE-NWM, but no primary contrast crosses GPU types. A planned secondary readout, the direct spread between A100 seeds, is available only for RAE-NWM (0.0824), because EgoWM's seeds were scored in separate sessions, and we form no A100 responsiveness ratio. The evidence is limited to RECON, 4\,s clips and injected faults; we make no claim about the alternative models' published scores or about downstream planning.
\end{document}